\documentclass{article}

\PassOptionsToPackage{numbers,compress}{natbib}
\usepackage[preprint]{neurips_2026}

\usepackage[utf8]{inputenc} % allow utf-8 input
\usepackage[T1]{fontenc}  % use 8-bit T1 fonts
\usepackage{hyperref}    % hyperlinks
\usepackage{url}      % simple URL typesetting
\usepackage{booktabs}    % professional-quality tables
\usepackage{amsfonts}    % blackboard math symbols
\usepackage{nicefrac}    % compact symbols for 1/2, etc.
\usepackage{microtype}   % microtypography
\usepackage[table]{xcolor}% colors
\usepackage{multirow}
\usepackage{siunitx}
\usepackage{graphicx}
\usepackage{tabularx}
\usepackage{amsmath}
\usepackage{array}
\usepackage{subcaption}
\usepackage{enumitem}
\usepackage{amssymb}
\usepackage{algorithm}
\usepackage{algpseudocode}
\usepackage{pifont}
\usepackage{arydshln}

\definecolor{vbenchbg}{RGB}{239,247,253}
\definecolor{vistbg}{RGB}{247,246,253}
\definecolor{narrabg}{RGB}{253,244,243}

\definecolor{scoreFirst}{RGB}{238,132,72}
\definecolor{scoreSecond}{RGB}{246,184,104}
\definecolor{scoreThird}{RGB}{250,223,171}

\newcolumntype{Y}{>{\centering\arraybackslash}X}

\newcommand{\third}[1]{#1}

\usepackage{newfloat}
\usepackage{listings}
\DeclareCaptionStyle{ruled}{labelfont=normalfont,labelsep=colon,strut=off} % DO NOT CHANGE THIS
\floatstyle{ruled}
\newfloat{listing}{tb}{lst}{}
\floatname{listing}{Listing}

\usepackage{booktabs}
\title{SlotMem: Character-Addressable Internal Memory for Narrative Long Video Generation}

\author{%
Yilai Liu\textsuperscript{1} \quad
Xin Zhang\textsuperscript{2} \quad
Shiyuan Zhang\textsuperscript{1} \quad
Hongyang Du\textsuperscript{1}\thanks{Corresponding author.} \\
\textsuperscript{1}The University of Hong Kong \\
\textsuperscript{2}Tsinghua University \\
\texttt{yilai\_liu@connect.hku.hk} \quad
\texttt{duhy@hku.hk}
}

\begin{document}

\maketitle


\begin{abstract}
Maintaining recurring character identities across scene transitions and long temporal gaps is a central challenge in narrative long video generation. Methods targeting global consistency often retrieve memory using cues that are not aligned with character identity preservation, while recent character-centric variants still rely on coarse frame-level kv memory that entangles identity with incidental visual factors and lacks a continuous update mechanism under limited memory capacity. To address these limitations, we propose \textbf{SlotMem}, a character-addressable internal memory framework for multi-character narrative long video generation. Specifically, SlotMem uses a Character-Semantic Probe to localize character-relevant visual tokens from cross-attention responses, and a Memory Encoder to compress DiT tokens into compact role-wise slot memory. As generation proceeds, a Memory Writer conservatively updates each character's memory with new observations, while Character-Wise Cross-Attention retrieves the role memory and injects it only into localized tokens of the same character. Experiments on multiple narrative long video generation benchmarks show that SlotMem improves long-range character consistency over existing baselines, while maintaining comparable video quality. Our code is available at \href{https://github.com/YilaiLiu-HKU/SlotMem}{https://github.com/YilaiLiu-HKU/SlotMem}.
% Our codes are available at: \url{https://anonymous.4open.science/r/JigsawMem}.
\end{abstract}


\section{Introduction}
Recent video generation models have made substantial progress in producing visually realistic short videos~\cite{seedance2,kling}. As generation moves from short videos to long narratives, the key challenge shifts from local visual fidelity to long-range consistency, especially character consistency.
This becomes especially challenging when characters disappear, reappear, change viewpoints, or interact with different scenes over extended durations. In practice, long video generation is commonly formulated as autoregressive video generation, where a long sequence is generated chunk by chunk with only a local historical context to fit within GPU memory constraints~\cite{selfforcing,selfforcingpp,svi}.
Although this strategy improves scalability, it constrains the model to a partial view of the generated history. As a result, previously observed features and appearance details may be discarded from the local context. When such information is needed again after a long temporal gap, the model may fail to reconstruct same character with consistent identity~\cite{svi}, leading to \textbf{identity drift}.

Existing methods reduce different forms of drift in long video generation through memory retrieval with task-specific criteria, such as geometric overlap, aesthetic quality, or semantic relevance to the upcoming chunk~\cite{contextasmemory,storymem,memflow,moc}. However, these criteria are not explicitly optimized for preserving recurring character identities across long temporal gaps. As a result, the retrieved memory may contain useful visual context, but it does not guarantee character-specific identity guidance and can only indirectly mitigate identity drift.

\begin{figure*}[!ht] % [!ht] 是一个可选参数,让LaTeX尽量把图片放在这里 (Here) 或顶部 (Top)
 \centering
\includegraphics[width=\linewidth]{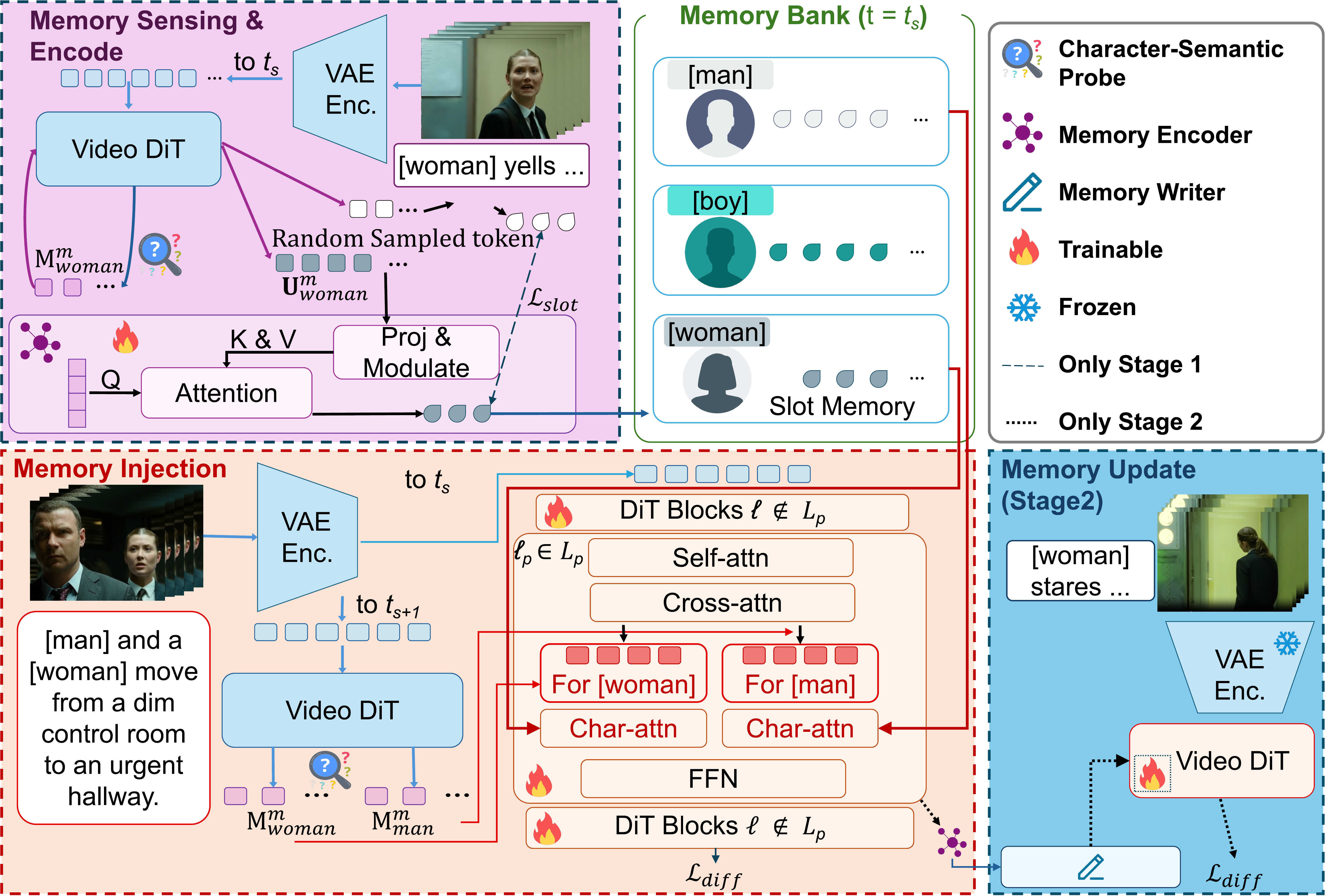}
\caption{Overview of \textbf{SlotMem}. It senses and encodes role-wise memory from the source chunk through Character-Semantic Probe and Memory Encoder (top-left), injects the memory through Character-Wise Cross-Attention (bottom-left), and updates the memory across the autoregressive generation process through the Memory Writer (bottom-right).}
 \label{fig:main} % 整个图的引用标签
 \vspace{-2em}
\end{figure*}
The recent identity-aware memory method IAMFlow~\cite{IAMFlow} takes an important step toward character-centric memory by using VLM scores to select character keyframes and incorporating their KV caches into generation. However, frame-level memory remains a holistic representation of the selected keyframe, where character identity is entangled with pose, viewpoint, background, lighting, and nearby visual elements. Reusing such frame-level features or KV caches may therefore transfer these incidental factors together with the desired identity cues, making the memory less flexible when the current chunk requires new poses, views, or interactions. Moreover, when the memory buffer becomes full, keyframe-based memory can only discard frames according to heuristic criteria such as low importance scores or temporal distance, rather than continuously updating the stored character memory with new observations.
These issues suggest that character memory should be better disentangled from backgrounds and other characters, while being dynamically updated by integrating previous memory with new observations.

This motivates us to introduce SlotMem, a character-addressable internal memory framework for narrative long video generation. Rather than storing selected keyframes as isolated historical observations, SlotMem represents each recurring character as a compact memory state extracted from feature tokens. Under a fixed memory budget, newly observed character features during generation are not appended as additional frame-level memory; instead, it is consolidated into the existing character slots through conservative updates. This changes memory management from discrete keyframe selection to internal memory extraction and evolution.

Built upon WAN2.2~\cite{wan}, SlotMem first uses a Character-Semantic Probe to sense relevant visual tokens for specific characters from cross-attention responses, and then encode their internal DiT features into role-wise slot memory through a Memory Encoder. During subsequent chunk generation, the corresponding character memory is retrieved and injected only into localized tokens of the same character through Character-Wise Cross-Attention. As generation proceeds, a Memory Writer updates the stored slots with reliable new observations while preserving the learned memory. This design reduces the entanglement between character identity and incidental visual factors such as background and co-occurring characters, enabling fine-grained identity guidance for narrative video generation.

% To address \textbf{character-centric disentanglement}, JigsawMem uses a Character-Semantic Probe to localize character-relevant visual tokens by measuring cross-attention responses between visual tokens and character name tokens. The localized features are then stored in role-wise memory banks, allowing each character's identity memory to be separated from backgrounds, scene states, and other characters.
% To achieve \textbf{model-native compatibility}, JigsawMem further identifies an identity-sensitive intermediate layer, where features are particularly informative for character identity, echoing the layer-wise geometric sensitivity observed in CAMEO~\cite{cameo}. During generation, features at the selected token positions and the identity-sensitive layer are stored in the corresponding role-wise memory bank. When the same character reappears, the stored memory is retrieved and injected into the current latent through Character-Wise Cross-Attention.
% By retrieving character-specific internal features only at localized token positions and injecting them at the identity-sensitive layer, JigsawMem provides fine-grained identity guidance while avoiding mismatches from external feature spaces or static references. In practice, it introduces only a lightweight character attention module at one intermediate layer, leaving the WAN2.2 pipeline otherwise unchanged and avoiding the overhead of global or multi-layer memory conditioning.
The main contributions are as follows:
\begin{itemize}[leftmargin=*]
% \item We formulate identity drift in narrative long video generation as a character-addressable internal memory problem, distinct from coarse memory retrieval and external reference conditioning.
\item We propose SlotMem, a role-wise memory framework that represents each recurring character with memory slots instead of keyframe memory, reducing entanglement with background, pose, and other characters.

\item We design a character-centric memory sensing, update, and injection mechanism. SlotMem localizes character-relevant tokens with a Character-Semantic Probe, compresses them into slot memory with a Memory Encoder, consolidates new observations through a Memory Writer, and injects the memory only into localized tokens of the same character.
\item We evaluate SlotMem on multi-character narrative long video generation using both general video quality benchmarks and identity similarity metrics, showing improved character consistency over existing baselines while maintaining comparable video quality.
\end{itemize}

\section{Related Work}
\paragraph{Long Video Generation.}
Long video generation remains challenging because models must preserve temporal coherence under limited context length and accumulated generation errors. Existing methods mainly address this problem by expanding context, compressing historical information, or stabilizing autoregressive rollout. LCT~\cite{lct} extends pretrained DiTs with dense long-context attention, while StreamingT2V~\cite{streamingt2v} and HoloCine~\cite{holocine} support long-form or controllable generation with explicit temporal modeling. Other methods reduce history cost through compact or recurrent context modeling, including FramePack~\cite{framepack}, PFP~\cite{pfp}, VideoSSM~\cite{videossm}, and OneStory~\cite{onestory}. Forcing-based methods mitigate train--test mismatch in autoregressive video diffusion~\cite{selfforcing,selfforcingpp,rollingforcing,sparseforcing}, and Stable Video Infinity~\cite{svi} extends this idea to cross-chunk autoregressive generation with error-recycling fine-tuning. These methods improve scalability and temporal stability, but mainly preserve frame-, chunk-, or rollout-level continuity, leaving recurring characters prone to identity drift after long gaps.

\paragraph{Memory-based Narrative Long Video Generation.}
Memory-based video generation improves long-range consistency by reusing earlier generated content. StoryDiffusion~\cite{storydiffusion} introduces consistent self-attention to improve long-range visual coherence, Context-as-Memory~\cite{contextasmemory} retrieves historical frames as reusable context for scene-consistent generation, StoryMem~\cite{storymem} reformulates multi-shot storytelling as iterative shot synthesis conditioned on an explicit visual memory bank, and MemFlow~\cite{memflow} dynamically retrieves relevant historical frames and activates only the most useful memory tokens for each upcoming chunk. MoC~\cite{moc} further formulates long video generation as internal information retrieval, where each query selects relevant context chunks through learned sparse routing. In addition, multi-agent systems such as MovieAgent~\cite{movieagent}, DreamFactory~\cite{dreamfactory}, AniMaker~\cite{animaker}, and VideoMemory~\cite{videomemory} leverage external memory and agent workflows to organize long-form generation based on short-video models. Although these methods demonstrate the value of memory for global visual consistency, such as backgrounds and style, their retrieval and memory injection are not character-centric. This causes target-character memory to be entangled with backgrounds, scene states, and other characters. IAMFlow~\cite{IAMFlow} builds character-centric memory from selected keyframes, but keyframe memory still causes two limitations: character identity remains entangled with incidental visual factors, and the memory can only be refreshed by replacing keyframes rather than continuously updating stored character representations with new observations.

\section{Methodology}
SlotMem addresses identity drift in autoregressive video generation by storing internal features of recurring characters and reusing them when the same character reappears. The framework consists of four components: (i) memory sensing with a \textbf{Character-Semantic Probe}, which identifies character-relevant visual tokens; (ii) memory extraction with a \textbf{Memory Encoder}, which compresses the selected internal features into compact slot memory; (iii) memory update with a \textbf{Memory Writer}, which aligns memory extracted from the current chunk with the stored memory and updates it through conservative residual corrections; and (iv) memory injection with \textbf{Character-Wise Cross-Attention}, which retrieves the corresponding role memory and injects it only into target tokens localized by the Character-Semantic Probe.

\paragraph{Autoregressive Video Generation with Memory.}
Given a sequence of chunk prompts \(\{\mathbf{p}^{k}\}_{k=1}^{K}\) and generated video chunks \(\{\mathbf{x}^{k}\}_{k=1}^{K}\), autoregressive video generation with memory retrieval can be factorized as
\begin{equation}
p_{\theta}(\mathbf{x}^{1:K}\mid \mathbf{p}^{1:K})
=
\prod_{k=1}^{K}
p_{\theta}\!\left(
\mathbf{x}^{k}
\mid
\mathbf{L}^{k},
\mathcal{R}(\mathcal{M}^{k-1}\mid \mathbf{q}^{k}),
\mathbf{p}^{k}
\right),
\label{eq:formul}
\end{equation}
where \(\mathbf{L}\), \(\mathcal{M}\), and \(\mathbf{q}\) denote the local context, memory, and retrieval cue, respectively.

Existing methods mainly differ in how they define \(\mathcal{M}\), \(\mathbf{q}\), and the retrieval rule \(\mathcal{R}\). Most methods focus on global consistency instead of character consistency and rely on cues \(\mathbf{q}\) like temporal positions, aesthetic quality, frame similarity, or prompt alignment~\cite{moc,storymem,memflow}, none of which are character-addressable. Thus, retrieved memory is not explicitly bound to a specific recurring character. On the other hand, identity-aware methods such as IAMFlow~\cite{IAMFlow} organize memory \(\mathcal{M}\) by character-wise importance but still retrieve frames as memory. Such coarse memory entangles the target character identity with others and the background, leading to feature entanglement.

SlotMem uses character semantics to retrieve character-relevant feature tokens, encodes them into role-wise slot memory, and stores them in the memory bank. Later, the corresponding slots are updated and injected only at localized character tokens. Such a mechanism can build purer memory and reduce interference from irrelevant characters.

\subsection{Character-Semantic Probe}
\label{section:sensing}
To construct character-addressable memory, we first locate the visual tokens associated with each character. We introduce the Character-Semantic Probe, which estimates important regions for the target character from text-to-visual cross-attention responses. The probe does not introduce additional supervision or external detectors, and the resulting character masks are used only to guide memory extraction and injection.

For a selected chunk \(\mathbf{x}^{i}\), we sample noise \(\boldsymbol{\epsilon}^{i}\) and forward the
chunk to timestep \(t_s\),
\begin{equation}
\mathbf{z}^{i}_{t_s}=q_{t_s}(\mathbf{x}^{i},\boldsymbol{\epsilon}^{i}),
\end{equation}
where \(q_{t_s}(\cdot)\) denotes the forward noising process defined by the diffusion scheduler. We feed \(\mathbf{z}^{i}_{t_s}\), the image condition \(\mathbf{I}^{i}\), and the prompt \(\mathbf{p}^{i}\) into the Video DiT, and cache the cross-attention responses between visual tokens and text tokens.

For each character \(c\in\mathcal{C}^{i}\) in the \(i\)-th chunk, let \(\mathcal{T}^{i}_{c}\) denote the text-token positions corresponding to the character name \(c\) in prompt \(\mathbf{p}^{i}\). Let
\(\mathbf{R}^{i}_{\ell,h}(t_s)\in\mathbb{R}^{F_i\times H_p\times W_p\times |\mathbf{p}^{i}|}\)
denote the text-to-visual cross-attention response at probe layer \(\ell\) and attention head \(h\). We define the character-semantic response as
\begin{equation}
\mathbf{A}^{i}_{c}(t_s)
=
\frac{1}{|\mathcal{L}_{p}|H|\mathcal{T}^{i}_{c}|}
\sum_{\ell\in\mathcal{L}_{p}}
\sum_{h=1}^{H}
\sum_{r\in\mathcal{T}^{i}_{c}}
\mathbf{R}^{i}_{\ell,h}(t_s)[:,:,:,r],
\label{eq:char_response}
\end{equation}
Thus, \(\mathbf{A}^{i}_{c}(t_s)\in\mathbb{R}^{F_i\times H_p\times W_p}\), and \(\mathbf{A}^{i}_{c}(t_s;f,u,v)\) scores how strongly the latent spatio-temporal patch at latent frame \(f\) and patch location \((u,v)\) responds to character \(c\).

When multiple characters appear in the same chunk, semantically similar character descriptions may produce overlapping cross-attention responses, causing identity interference between roles. To prevent such interference, we subtract the average response of the other characters in the same chunk.
\begin{equation}\label{eq:diff_response}
\hat{\mathbf{A}}^{i}_{c}(t_s)
=
\begin{cases}
\mathbf{A}^{i}_{c}(t_s), & |\mathcal{C}^{i}|=1, \\[4pt]
\mathbf{A}^{i}_{c}(t_s)
-
\frac{1}{|\mathcal{C}^{i}|-1}
\sum\limits_{c'\in\mathcal{C}^{i}\setminus\{c\}}
\mathbf{A}^{i}_{c'}(t_s), & |\mathcal{C}^{i}|>1,
\end{cases}
\end{equation}
We then select character-sensitive tokens from the normalized differential response. We apply top-\(p\) selection and remove isolated noisy tokens with a local continuity filter. The character mask is defined as
\begin{equation}
\mathbf{M}^{i}_{c}(t_s)
=
\operatorname{Cont}_{\kappa}\!\left(
\operatorname{Top}_{p}\!\left(\hat{\mathbf{A}}^{i}_{c}(t_s)\right)
\right),
\label{eq:char_mask}
\end{equation}
Here, \(\operatorname{Top}_{p}(\cdot)\) keeps the top-\(p\) fraction of latent visual tokens, and \(\operatorname{Cont}_{\kappa}(\cdot)\) removes isolated selections using a local spatial-neighborhood filter on each latent frame.

The resulting mask \(\mathbf{M}^{i}_{c}(t_s)\) localizes the tokens most relevant to the semantics of character \(c\) in chunk \(i\). It provides a character-specific interface for memory extraction and update, allowing SlotMem to extract relevant internal features while reducing interference from background and other visual noise. During memory injection, the same mask constrains the retrieved memory to relevant tokens, avoiding unnecessary modifications to irrelevant regions.

\subsection{Character-Addressable Memory}
\label{Mem}

Given a target chunk \(\mathbf{x}^{k}\) with prompt \(\mathbf{p}^{k}\), SlotMem constructs character-addressable memory from previous chunks in the same video. A memory chunk \(\mathbf{x}^{m}\) provides the initial memory, while an update chunk \(\mathbf{x}^{u}\) provides evidence for refinement. Their prompts are denoted by \(\mathbf{p}^{m}\), \(\mathbf{p}^{u}\), and \(\mathbf{p}^{k}\).

Memory extraction targets characters shared by the memory and target chunks,
\begin{equation}
\mathcal{S}_{\mathrm{ext}}^{m,k}
=
\mathcal{C}^{m}\cap\mathcal{C}^{k},
\end{equation}
where \(\mathcal{C}^{i}\) denotes the character names in prompt \(\mathbf{p}^{i}\). Memory update is applied only when the same character is also observed in the update chunk,
\begin{equation}
\mathcal{S}_{\mathrm{upd}}^{m,u,k}
=
\mathcal{C}^{m}\cap\mathcal{C}^{u}\cap\mathcal{C}^{k}.
\end{equation}
Accordingly, SlotMem is trained in two stages: the first stage learns memory extraction and injection, while the second stage learns memory update. Characters that do not occur in \(\mathcal{C}^{u}\)  keep their extracted memory unchanged.

\subsubsection{Memory Extraction with Memory Encoder}
\label{Mem Ext}

For memory extraction, the memory chunk is processed at the same timestep as the current denoising forward pass. Specifically, we apply the Character-Semantic Probe to the memory chunk \(\mathbf{x}^{m}\) with prompt \(\mathbf{p}^{m}\) at timestep \(t_s\), producing a character mask \(\mathbf{M}^{m}_{c}(t_s)\). For each selected DiT layer \(\ell\in\mathcal{L}_{p}\), we collect the localized internal tokens for specific characters,
\begin{equation}
\mathbf{U}^{m,\ell}_{c}(t_s)
=
\mathbf{h}^{m}_{\ell}(t_s)[\mathbf{M}^{m}_{c}(t_s)]
\in\mathbb{R}^{n_{c,\ell}(t_s)\times d}.
\label{eq:Mem exe probe}
\end{equation}

These tokens are already at the same denoising stage as the target forward pass. We use them as the memory source and feed them into the Memory Encoder. The encoder first projects the localized tokens into a bottleneck space and modulates them with the timestep embedding,
\begin{equation}
\widetilde{\mathbf{G}}^{m,\ell}_{c}(t_s)
=
\phi_{t_s}\!\left(
\operatorname{LN}(\mathbf{U}^{m,\ell}_{c}(t_s))\mathbf{W}_{\mathrm{in}}
\right),
\label{mem exe gate}
\end{equation}
where \(\phi_{t_s}\) denotes standard timestep affine modulation. The encoder then uses \(R\) learned slot queries to pool the adapted role tokens,
\begin{equation}
\mathbf{P}^{m,\ell}_{c}(t_s)
=
\operatorname{softmax}
\left(
\frac{
\mathbf{Q}_{g(\ell)}
(\widetilde{\mathbf{G}}^{m,\ell}_{c}(t_s))^{\top}
}{\sqrt{d_e}}
\right)
\widetilde{\mathbf{G}}^{m,\ell}_{c}(t_s).
\label{mem exe pool}
\end{equation}
The FFN projects the pooled slots back to the DiT feature dimension, and \(\mathbf{P}_{\mathrm{slot}}\) is a learned slot-index embedding that distinguishes different memory slots:
\begin{equation}
\mathbf{S}^{m,\ell}_{c}(t_s)
=
\operatorname{FFN}_{g(\ell)}
(\mathbf{P}^{m,\ell}_{c}(t_s))
+
\mathbf{P}_{\mathrm{slot}}.
\label{mem exe encode}
\end{equation}

\subsubsection{Memory Update with Memory Writer}
\label{Mem Update}

The memory extracted from a single chunk may miss later observations of the same character. To make the memory dynamic, we introduce a Memory Writer that refines the stored slots with evidence from an update chunk. The update is performed independently for each character, so memory from different roles is not mixed.

Let \(\mathbf{s}^{m,\ell}_{c,j}(t_s)\) be the \(j\)-th stored slot of character \(c\), and let \(\mathbf{S}^{u,\ell}_{c}(t_s)\) denote the new memory extracted by the Memory Encoder from the update chunk \(\mathbf{x}^{u}\) .
For each stored slot, the writer retrieves an update context from the update slots
\begin{equation}
\mathbf{c}^{u,\ell}_{c,j}(t_s)
=
\operatorname{Attn}
\left(
\operatorname{LN}(\mathbf{s}^{m,\ell}_{c,j}(t_s)),
\operatorname{LN}(\mathbf{S}^{u,\ell}_{c}(t_s)),
\mathbf{S}^{u,\ell}_{c}(t_s)
\right).
\label{mem up context}
\end{equation}

Based on the stored slot and the retrieved context, the writer predicts a residual correction,
\begin{equation}
\Delta^{\ell}_{c,j}(t_s)
=
\operatorname{MLP}_{\Delta}
\left(
[
\operatorname{LN}(\mathbf{s}^{m,\ell}_{c,j}(t_s));
\operatorname{LN}(\mathbf{c}^{u,\ell}_{c,j}(t_s))
]
\right).
\label{mem up delta}
\end{equation}

To avoid unreliable updates, we add a gate control using the cosine similarity between the stored slot and the observation context to decide whether the update should be applied, and an MLP to predict a learnable update scale
\begin{equation}
g^{\ell}_{c,j}(t_s)
=
\mathbb{I}
\left[
\operatorname{cos}
\left(
\mathbf{s}^{m,\ell}_{c,j}(t_s),
\mathbf{c}^{u,\ell}_{c,j}(t_s)
\right)
>
\tau_{\mathrm{prec}}
\right]
\cdot
\sigma
\left(
\operatorname{MLP}_{g}
\left(
[
\operatorname{LN}(\mathbf{s}^{m,\ell}_{c,j}(t_s));
\operatorname{LN}(\mathbf{c}^{u,\ell}_{c,j}(t_s))
]
\right)
\right).
\label{mem up gate}
\end{equation}

The stored slot is updated through a gated residual correction
\begin{equation}
\widetilde{\mathbf{s}}^{m,\ell}_{c,j}(t_s)
=
\mathbf{s}^{m,\ell}_{c,j}(t_s)
+
g^{\ell}_{c,j}(t_s)\,
\Delta^{\ell}_{c,j}(t_s),
\label{mem up}
\end{equation}

where \(\tau_{\mathrm{prec}}\) is the similarity threshold. Characters without reliable update evidence keep their stored slots unchanged.

\begin{table*}[t]
\centering
\small
\setlength{\tabcolsep}{2.6pt}
\renewcommand{\arraystretch}{1.02}
\begin{tabularx}{\linewidth}{@{}>{\raggedright\arraybackslash}p{0.23\textwidth}*{5}{Y}@{}}
\toprule
\textbf{Metric \(\uparrow\)}
& \textbf{Wan2.2-I2V}
& \textbf{+StoryDiff.}
& \textbf{+StoryMem}
& \textbf{+IAMFlow}
& \textbf{+SlotMem} \\
\midrule

\multicolumn{6}{c}{\textbf{VBench}} \\
\midrule
Background Consistency & 0.8580 & \textbf{0.8950} & 0.8659 & \underline{0.8929} & \third{0.8832} \\
Motion Smoothness      & \third{0.9850} & 0.9805 & 0.9849 & \underline{0.9908} & \textbf{0.9912} \\
Dynamic Degree         & 0.7826 & \underline{0.8696} & \textbf{0.9130} & 0.3913 & \underline{0.8696} \\
Aesthetic Quality      & \underline{0.5688} & \textbf{0.6091} & 0.5260 & 0.5628 & \third{0.5651} \\
Imaging Quality        & \third{0.6154} & \underline{0.6620} & 0.5618 & \textbf{0.7119} & 0.5963 \\
Human Anatomy          & \third{0.9381} & \underline{0.9440} & 0.8892 & 0.9062 & \textbf{0.9480} \\
\midrule

\multicolumn{6}{c}{\textbf{ViStoryBench}} \\
\midrule
Style Similarity       & 0.7279 & \third{0.8123} & 0.8040 & \underline{0.8649} & \textbf{0.8819} \\
Character Similarity   & 0.7701 & \third{0.8098} & 0.7592 & \underline{0.8446} & \textbf{0.8603} \\
Prompt Alignment       & \underline{0.8299} & 0.7627 & \third{0.8273} & 0.7192 & \textbf{0.8733} \\
Character Matching     & \third{0.9877} & \underline{0.9879} & 0.9789 & 0.9859 & \textbf{0.9957} \\
Inception Score        & \third{5.0985} & \textbf{6.6777} & 4.7509 & 4.8609 & \underline{6.0769} \\
Aesthetic Score        & \third{4.6121} & \textbf{5.3475} & 4.3363 & \underline{4.7729} & 4.5278 \\
Copy-Paste (Comp.)& \third{0.4789} & 0.4689 & \underline{0.5004} & 0.4692 & \textbf{0.5631} \\
\midrule

\multicolumn{6}{c}{\textbf{NarraStream-Bench}} \\
\midrule
Subject Consistency             & \third{0.8427} & 0.6927 & 0.7181 & \underline{0.8524} & \textbf{0.8771} \\
Background Consistency          & 0.8135 & 0.8146 & \third{0.8339} & \textbf{0.8485} & \underline{0.8444} \\
Motion Smoothness               & \third{0.3442} & 0.3186 & 0.3360 & \underline{0.4681} & \textbf{0.5166} \\
Temporal Flickering             & \third{0.7523} & 0.6204 & 0.6933 & \textbf{0.8614} & \underline{0.8181} \\
Video Temporal Stability  & 0.5859 & 0.4133 & \third{0.6272} & \textbf{0.7527} & \underline{0.6836} \\
Boundary Smoothness             & 0.3102 & 0.1778 & \underline{0.5231} & \third{0.4368} & \textbf{0.7529} \\
Conditional Adjacent            & 0.3632 & 0.3264 & \third{0.4556} & \textbf{0.6131} & \underline{0.6126} \\
Conditional Long-range          & \third{0.7239} & 0.5880 & 0.6165 & \underline{0.7313} & \textbf{0.8363} \\
Dynamic Trajectory              & 0.4550 & \third{0.4935} & \underline{0.5235} & \textbf{0.5533} & 0.4723 \\
Entity Grounding                & \third{0.6633} & 0.5675 & \underline{0.6648} & 0.6029 & \textbf{0.6735} \\
VLM Score                       & \third{0.4708} & 0.2817 & \underline{0.4735} & 0.4607 & \textbf{0.5384} \\
\bottomrule
\end{tabularx}

\caption{
Performance comparison of the baselines and SlotMem across different benchmarks.
All models use Wan2.2-I2V-A14B as the backbone.
}
\label{tab:three_bench_results_and_radars}
\end{table*}

\subsection{Memory Injection with Character-Wise Cross-Attention}
\label{Mem Inj}
Given memory slots at timestep \(t_s\), SlotMem injects identity information only into target tokens that correspond to the same character. During denoising, the injection positions are localized by the mask from the previous step. The injection positions are given by the mask estimated in the previous forward step
\begin{equation}
\mathbf{M}^{k,\ell}_{c}(t_{s-1})
=
\operatorname{Probe}_{\ell}
\left(
\mathbf{z}^{k}_{t_{s-1}},
\mathbf{I}^{k},
\mathbf{p}^{k},
c
\right).
\label{mem in probe}
\end{equation}
Thus, the mask is lagged by one step, while the injected memory and target features are taken from the current step.

Let \(\bar{\mathbf{S}}^{\ell}_{c}(t_s)\in\mathbb{R}^{R\times d}\) denote the current slot memory of character \(c\) at injection layer \(\ell\in\mathcal{L}_{p}\). The relevant tokens are extracted from the hidden states at step \(t_s\) using the mask at the previous timestep
\begin{equation}
\mathbf{Q}^{k,\ell}_{c}(t_s)
=
\mathbf{h}^{k}_{\ell}(t_s)
\left[
\mathbf{M}^{k,\ell}_{c}(t_{s-1})
\right],
\qquad
c\in\mathcal{S}_{\mathrm{ext}}^{m,k}.
\label{eq:query_extract}
\end{equation}

At each layer \(\ell\), SlotMem performs Character-Wise Cross-Attention. The current character tokens serve as queries, and the current memory slots serve as keys and values:
\begin{equation}
\Delta \mathbf{Q}^{k,\ell}_{c}(t_s)
=
\operatorname{Attn}\!\left(
\mathbf{Q}^{k,\ell}_{c}(t_s)\mathbf{W}^{\ell}_{Q},
\bar{\mathbf{S}}^{\ell}_{c}(t_s)\mathbf{W}^{\ell}_{K},
\bar{\mathbf{S}}^{\ell}_{c}(t_s)\mathbf{W}^{\ell}_{V}
\right).
\label{eq:char_cross_attn}
\end{equation}
The attended memory is projected back and fused with the localized character positions:
\begin{equation}
\mathbf{h}^{k}_{\ell}(t_s)
\left[
\mathbf{M}^{k,\ell}_{c}(t_{s-1})
\right]
=
\mathbf{h}^{k}_{\ell}(t_s)
\left[
\mathbf{M}^{k,\ell}_{c}(t_{s-1})
\right]
+
\alpha_{\ell}(t_s)\mathbf{W}^{\ell}_{O}
\Delta \mathbf{Q}^{k,\ell}_{c}(t_s).
\label{eq:sparse_writeback}
\end{equation}
Here, \(\alpha_{\ell}(t_s)\) is a learnable timestep gate that controls the strength of memory injection at layer \(\ell\). This sparse injection keeps identity injection local to the corresponding character, reducing interference with background and other characters.

\begin{figure*}[!t] 
 \centering
\includegraphics[width=\textwidth]{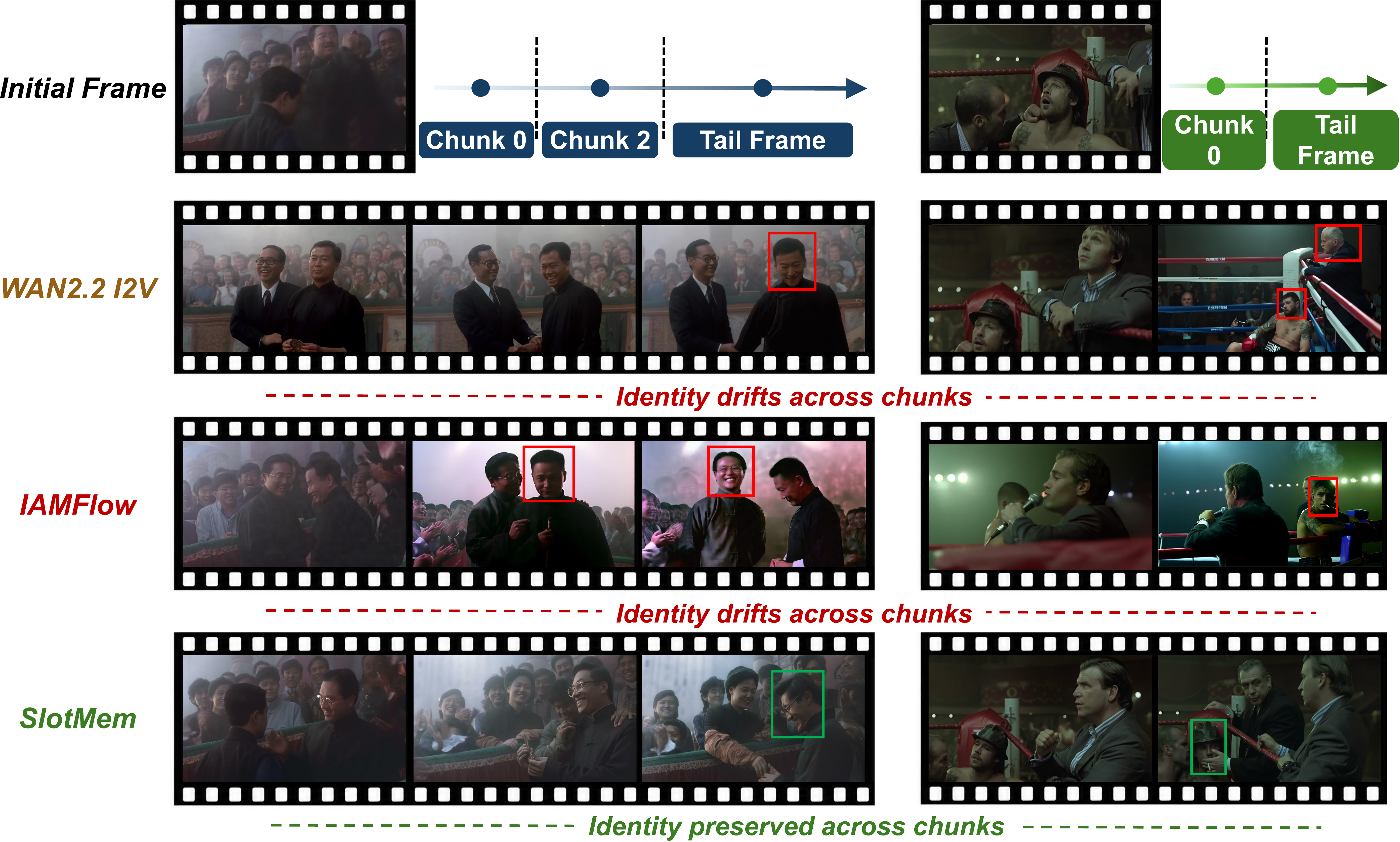}
\caption{
Qualitative comparison of the final frames generated for corresponding chunks by different methods. From left to right, each column presents the tail frame of the same generated chunk produced by each baseline and by SlotMem. Red bounding boxes indicate regions where the baselines fail to preserve character identity or produce unstable facial details, while green bounding boxes highlight the stronger ability of SlotMem to maintain character consistency across chunks.
}
 \label{fig:puzzle_visual} % 整个图的引用标签
\end{figure*}

\subsection{Training Stage}

SlotMem is trained in two stages to decouple memory construction from memory update. In the first stage, each training sample contains a memory chunk \(\mathbf{x}^{m}\) and a target chunk \(\mathbf{x}^{k}\). The memory chunk provides character-specific source features, which are encoded into slot memory by the Memory Encoder and injected into the target chunk through Character-Wise Cross-Attention. During this stage, we optimize the Memory Encoder and Character-Wise Cross-Attention, and fine-tune the DiT backbone.

To make the slots discriminative across characters and separable from irrelevant regions, we add an auxiliary contrastive loss. Character slots are obtained from probe-selected role tokens, while background slots are obtained by randomly sampling tokens outside all character masks and passing them through the same Memory Encoder. For each character or background class, we compute the mean feature of its encoded slots as a class prototype. Each slot is then classified by its cosine similarity to all class prototypes
\begin{equation}
\mathcal{L}_{\mathrm{slot}}
=
-\frac{1}{|\mathcal{J}|}
\sum_{j\in\mathcal{J}}
\log
\frac{
\exp(\mathbf{f}_{j}(t_s)^{\top}\boldsymbol{\mu}_{y_j}(t_s)/\tau_{\mathrm{slot}})
}{
\sum_{\rho}
\exp(\mathbf{f}_{j}(t_s)^{\top}\boldsymbol{\mu}_{\rho}(t_s)/\tau_{\mathrm{slot}})
},
\label{mem exe contra}
\end{equation}
where \(y_j\) is the role/background label of slot \(j\), \(\mathcal{J}\) is the set of encoded character and background slots, and \(\tau_{\mathrm{slot}}\) is the temperature. The first-stage training objective is
\begin{equation}
\mathcal{L}_{\mathrm{stage1}}
=
\mathcal{L}_{\mathrm{diff}}
+
\lambda_{\mathrm{slot}} \mathcal{L}_{\mathrm{slot}}
\label{mem exe loss}
\end{equation}
Here, \(\mathcal{L}_{\mathrm{diff}}\) denotes the standard denoising objective of the base video diffusion model, applied with the corresponding conditioning inputs and memory modules when enabled.

In the second stage, we introduce an update chunk \(\mathbf{x}^{u}\) to train dynamic memory refinement. The Memory Encoder and Character-Wise Cross-Attention modules are frozen, and the update chunk provides additional observations of recurring characters. The Memory Writer learns to align the stored memory slots with the update slots and predicts conservative residual corrections. During this stage, we optimize the Memory Writer and further fine-tune the modules in the first stage and DiT backbone with the standard denoising objective \(\mathcal{L}_{\mathrm{diff}}\).

This two-stage training strategy first learns to build and inject character-addressable memory, and then learns to update the stored memory without disrupting the learned memory interface. The full pipeline is illustrated in Fig.~\ref{fig:main}.

\subsection{Inference Stage}
At inference time, SlotMem runs online with the normal autoregressive denoising process and does not introduce an additional probing forward at the last timestep. In each DiT forward pass, the intermediate cross-attention responses are used by the Character-Semantic Probe to obtain character masks, and the visual features selected by these masks are used to construct or update memory. The obtained masks guide memory extraction and update for the current chunk, and are cached as the injection masks for the next denoising step. Therefore, memory operations are performed within the original generation process without adding extra inference forwards.

\section{Experiments}
\subsection{Experimental Settings}
\subsubsection{Data Curation}
\label{data curation}
We curate publicly available films and videos with high ratings from online movie rating platforms following the general practice of MultiShotMaster~\cite{multishotmaster}, which annotate each group of video chunks with a hierarchical captioning strategy. For each group, we first generate a global caption that describes the main characters and the overall scene. Unlike the ``Subject 1'' naming rule used in MultiShotMaster, we require the captioner to assign visually discriminative character names, such as ``standing apprentice'', ``blonde man'', or ``white car''. When generating each chunk-level prompt, the captioner is explicitly instructed to reuse the exact character names from the global caption whenever the corresponding character appears. This rule enforces name consistency across chunks and allows the names to serve as semantic anchors for recurring identities.

After manually filtering the curated data, we construct a character mapping from the captions. During training, this metadata is used to select valid memory--target pairs, where the memory chunk must precede the target chunk and contain at least one character name that also appears in the target chunk. This criterion ensures that memory injection is applied only when the target chunk contains a recurring character whose identity can be retrieved from the preceding context.

\subsubsection{Implementation Details.}
We build our framework upon the Wan2.2-I2V-A14B model~\cite{wan}. We fine-tune the model with rank-128 LoRA while keeping the pretrained backbone frozen. The model is trained on our curated dataset, where each training sample consists of a target chunk, its image condition, a preceding memory chunk and an update chunk that share at least one character name with the target chunk. Each chunk contains 81 frames, corresponding to 3.375 seconds at 24 FPS, and adjacent chunks are constructed with 5 overlapping frames to preserve local temporal continuity.

\subsubsection{Evaluation Metrics}

\paragraph{Visual Quality.}
We use \textbf{VBench}~\cite{vbench} to assess general video quality, including background consistency, motion smoothness, dynamic degree, aesthetic quality, and imaging quality. Since the original VBench subject consistency metric is unreliable in multi-character videos, where multiple identities may appear and disappear asynchronously, we replace it with the Human Anatomy from \textbf{VBench2}~\cite{vbench2} to measure human structural correctness.

\paragraph{Character and Story Consistency.}
We use \textbf{ViStoryBench}~\cite{vistorybench} and \textbf{NarraStream-Bench}~\cite{IAMFlow} to evaluate character and story consistency from complementary perspectives.
% Since curated data does not provide annotated character reference images required by \textbf{ViStoryBench}, we construct a fully automatic pipeline with VLM annotation and VFM detection to obtain pseudo reference images from the original captions and generated videos. The full pipeline is illustrated in Appendix~\ref{pseudo_pipe}.
%  These pseudo references are used for reference dependent character metrics such as CIDS. 
In the result tables, we report \(1-\mathrm{Copy\ Paste}\) instead of \(\mathrm{Copy\ Paste}\), so that all displayed scores are positively correlated with generation quality. For consistent evaluation, we use GPT-4.1 ~\cite{gpt41} as the evaluator for all VLM-based metrics in both benchmarks.

\subsection{Performance}

\paragraph{Quantitative Results.}

Table~\ref{tab:three_bench_results_and_radars} compares SlotMem with the
baselines in visual quality, character consistency, and narrative coherence.
Wan2.2 provides a strong generation baseline but remains limited on
character-related metrics. StoryDiffusion achieves competitive visual quality
on VBench but degrades substantially on NarraStream-Bench. StoryMem preserves
overall video dynamics, as reflected by its high Dynamic Degree and Dynamic
Trajectory scores, yet exhibits weak subject consistency. IAMFlow improves
subject consistency and temporal stability, but its high Motion Smoothness is
accompanied by a markedly lower VBench Dynamic Degree of 0.3913. Since Dynamic
Degree measures whether a video contains sufficiently large visible motion,
including both subject and camera motion, this result suggests that IAMFlow
often produces overly static videos.

In contrast, SlotMem achieves the best Human Anatomy score on VBench
(0.9480), Character Similarity on ViStoryBench (0.8603), and Subject
Consistency on NarraStream-Bench (0.8771). It also obtains the highest VBench
Motion Smoothness score of 0.9912 while maintaining a competitive Dynamic
Degree of 0.8696. These results demonstrate that SlotMem improves
recurring-character consistency and temporal stability without suppressing
the overall dynamics of the generated videos.

\paragraph{Qualitative Results.}
As shown in Fig.~\ref{fig:puzzle_visual}, the baselines often lose identity cues when a character reappears in later chunks. The red boxes highlight changes in facial structure and hair appearance, as well as unstable rendering under motion. The performance of IAMFlow is limited by identity entanglement in multi-character keyframes. In contrast, SlotMem retrieves role-wise internal features and injects them only at localized character tokens, leading to more stable identity cues across chunks and fewer motion-induced artifacts.

\begin{table}[!ht]
\centering
\small
\setlength{\tabcolsep}{3.0pt}
\renewcommand{\arraystretch}{1.06}

% Make tabularx X/Y columns vertically centered.
\renewcommand{\tabularxcolumn}[1]{m{#1}}

% Internal vertical padding inside colored cells.
\newcommand{\cellpad}[1]{%
  \rule[-1.05ex]{0pt}{3.7ex}#1%
}

% Compact two-line model name, with almost no gap between the two lines.
\newcommand{\modeltwoline}[2]{%
  \shortstack[l]{%
    \textbf{#1}\\
    \textbf{#2}%
  }%
}

\begin{tabularx}{\linewidth}
{@{}>{\raggedright\arraybackslash}m{0.22\linewidth}@{}*{4}{Y@{}}}
\toprule
\textbf{Model}
& \textbf{Char. Sim.}
& \textbf{Prompt Align.}
& \textbf{Char. Match}
& \textbf{Inception} \\
\midrule

\cellpad{\textbf{Wan2.2}}
& \cellpad{0.7701}
& \cellpad{0.8299}
& \cellpad{0.9877}
& \cellpad{\third{5.0985}} \\
\midrule

\cellpad{\textbf{SlotMem}{ w/o M Enc.}}
& \cellpad{0.7852} 
& \cellpad{0.8314}
& \cellpad{0.9820}
& \cellpad{5.3210} \\

\cellpad{\textbf{SlotMem}{ w/o M Upd.}}
& \cellpad{\underline{0.7965}} 
& \cellpad{\underline{0.8439}}
& \cellpad{\underline{0.9880}}
& \cellpad{\underline{5.5840}} \\
\midrule

\cellpad{\textbf{SlotMem Full}}
& \cellpad{\textbf{0.8603}}
& \cellpad{\textbf{0.8733}}
& \cellpad{\textbf{0.9957}}
& \cellpad{\textbf{6.0769}} \\
\bottomrule
\end{tabularx}

\caption{Ablation studies on the memory modules in SlotMem.}
\label{tab:vistory_ablation_table}
\end{table}
\vspace{-1em}
\subsection{Ablation Study}
We ablate the Memory Encoder and Memory Writer while keeping the Character-Semantic Probe and Character-Wise Cross-Attention unchanged. Removing the probe would require dense full-context memory, while removing Character-Wise Cross-Attention would disable memory injection and reduce the model to the native Wan2.2 baseline. We therefore evaluate two key variants: first, removing the Memory Encoder and using raw probe-selected DiT tokens as static memory, which also disables the Memory Writer since no slot memory is formed; second, removing only the Memory Writer while keeping the Memory Encoder, so that the model uses memory from the initial observations without updates.

As shown in Table~\ref{tab:vistory_ablation_table}, removing either component consistently underperforms the full model. The Memory Encoder improves role-wise memory by compressing raw localized tokens into cleaner slots, while the Memory Writer further strengthens long-range identity preservation by updating stored memory with new observations.

% \begin{table}[ht]
% \centering
% \small
% \setlength{\tabcolsep}{8pt}
% \renewcommand{\arraystretch}{1.12}

% \begin{tabular}{lcccc}
% \toprule
% \textbf{Model} 
% & \textbf{Human Anatomy} 
% & \textbf{DINO Feature Similarity} 
% & \textbf{CLIP Feature Similarity} \\
% \midrule

% Wan2.1-I2V
% & 0.9142
% & 0.5550
% & 0.5993 \\
% Wan2.2-I2V
% & 0
% & 0
% & 0 \\
% Wan2.2-I2V HighNoise Fine-tuned
% & 0
% & 0
% & 0 \\

% Wan2.2-I2V Both Fine-tuned
% & 0
% & 0
% & 0 \\

% Wan2.2-I2V LowNoise Fine-tuned
% & 0
% & 0
% & 0 \\

% \bottomrule
% \end{tabular}

% \caption{
% Ablation study, add char cross-attn and fine-tuned on different part of the model.
% }
% \label{tab:ablation_key_metrics}
% \end{table}

\section{Conclusion}
We introduced SlotMem, a character-addressable internal memory framework for narrative long video generation. Instead of relying on coarse frame-level retrieval, SlotMem represents recurring characters with compact role-wise memory that can be updated across autoregressive video generation and injected only into character-relevant regions. Experiments on multi-character narrative long video generation demonstrate improved long-range identity consistency over existing baselines, while preserving comparable video quality.

\bibliography{refs}

@misc{seedance2,
      title={Seedance 2.0: Advancing Video Generation for World Complexity}, 
      author={{Team Seedance} and others},
      year={2026},
      eprint={2604.14148},
      archivePrefix={arXiv},
      primaryClass={cs.CV},
}

@misc{kling,
      title={Kling-Omni Technical Report}, 
      author={{Kling Team} and others},
      year={2025},
      eprint={2512.16776},
      archivePrefix={arXiv},
      primaryClass={cs.CV},
      url={https://arxiv.org/abs/2512.16776}, 
}

@misc{lct,
  title         = {Long Context Tuning for Video Generation},
  author        = {Yuwei Guo and Ceyuan Yang and Ziyan Yang and
                   Zhibei Ma and Zhijie Lin and Zhenheng Yang and
                   Dahua Lin and Lu Jiang},
  year          = {2025},
  eprint        = {2503.10589},
  archivePrefix = {arXiv},
  primaryClass  = {cs.CV},
  url           = {https://arxiv.org/abs/2503.10589}
}

@misc{holocine,
      title={HoloCine: Holistic Generation of Cinematic Multi-Shot Long Video Narratives}, 
      author={Yihao Meng and Hao Ouyang and Yue Yu and Qiuyu Wang and Wen Wang and Ka Leong Cheng and Hanlin Wang and Yixuan Li and Cheng Chen and Yanhong Zeng and Yujun Shen and Huamin Qu},
      year={2025},
      eprint={2510.20822},
      archivePrefix={arXiv},
      primaryClass={cs.CV},
      url={https://arxiv.org/abs/2510.20822}, 
}

@misc{storymem,
      title={StoryMem: Multi-shot Long Video Storytelling with Memory}, 
      author={Kaiwen Zhang and Liming Jiang and Angtian Wang and Jacob Zhiyuan Fang and Tiancheng Zhi and Qing Yan and Hao Kang and Xin Lu and Xingang Pan},
      year={2025},
      eprint={2512.19539},
      archivePrefix={arXiv},
      primaryClass={cs.CV},
      url={https://arxiv.org/abs/2512.19539}, 
}

@misc{memflow,
      title={MemFlow: Flowing Adaptive Memory for Consistent and Efficient Long Video Narratives}, 
      author={Sihui Ji and Xi Chen and Shuai Yang and Xin Tao and Pengfei Wan and Hengshuang Zhao},
      year={2025},
      eprint={2512.14699},
      archivePrefix={arXiv},
      primaryClass={cs.CV},
      url={https://arxiv.org/abs/2512.14699}, 
}

@misc{videossm,
      title={VideoSSM: Autoregressive Long Video Generation with Hybrid State-Space Memory}, 
      author={Yifei Yu and Xiaoshan Wu and Xinting Hu and Tao Hu and Yangtian Sun and Xiaoyang Lyu and Bo Wang and Lin Ma and Yuewen Ma and Zhongrui Wang and Xiaojuan Qi},
      year={2025},
      eprint={2512.04519},
      archivePrefix={arXiv},
      primaryClass={cs.CV},
      url={https://arxiv.org/abs/2512.04519}, 
}

@misc{svi,
      title={Stable Video Infinity: Infinite-Length Video Generation with Error Recycling}, 
      author={Wuyang Li and Wentao Pan and Po-Chien Luan and Yang Gao and Alexandre Alahi},
      year={2025},
      eprint={2510.09212},
      archivePrefix={arXiv},
      primaryClass={cs.CV},
      url={https://arxiv.org/abs/2510.09212}, 
}

@misc{selfforcing,
      title={Self Forcing: Bridging the Train-Test Gap in Autoregressive Video Diffusion}, 
      author={Xun Huang and Zhengqi Li and Guande He and Mingyuan Zhou and Eli Shechtman},
      year={2025},
      eprint={2506.08009},
      archivePrefix={arXiv},
      primaryClass={cs.CV},
      url={https://arxiv.org/abs/2506.08009}, 
}

@article{contextasmemory,
      title={Context as Memory: Scene-Consistent Interactive Long Video Generation with Memory Retrieval},
      author={Yu, Jiwen and Bai, Jianhong and Qin, Yiran and Liu, Quande and Wang, Xintao and Wan, Pengfei and Zhang, Di and Liu, Xihui},
      journal={arXiv preprint arXiv:2506.03141},
      year={2025}
}

@inproceedings{moc,
  title     = {Mixture of Contexts for Long Video Generation},
  author    = {Cai, Shengqu and Yang, Ceyuan and Zhang, Lvmin and Guo, Yuwei and Xiao, Junfei and Yang, Ziyan and Xu, Yinghao and Yang, Zhenheng and Yuille, Alan and Guibas, Leonidas and Agrawala, Maneesh and Jiang, Lu and Wetzstein, Gordon},
  booktitle = {ICLR},
  year      = {2026},
}

@misc{multishotmaster,
      title={MultiShotMaster: A Controllable Multi-Shot Video Generation Framework}, 
      author={Qinghe Wang and Xiaoyu Shi and Baolu Li and Weikang Bian and Quande Liu and Huchuan Lu and Xintao Wang and Pengfei Wan and Kun Gai and Xu Jia},
      year={2025},
      eprint={2512.03041},
      archivePrefix={arXiv},
      primaryClass={cs.CV},
      url={https://arxiv.org/abs/2512.03041}, 
}

@misc{wan,
      title={Wan: Open and Advanced Large-Scale Video Generative Models}, 
      author={{Wan Team} and others},
      year={2025},
      eprint={2503.20314},
      archivePrefix={arXiv},
      primaryClass={cs.CV},
      url={https://arxiv.org/abs/2503.20314}, 
}

@article{streamingt2v,
    title={StreamingT2V: Consistent, Dynamic, and Extendable Long Video Generation from Text},
    author={Henschel, Roberto and Khachatryan, Levon and Hayrapetyan, Daniil and Poghosyan, Hayk and Tadevosyan, Vahram and Wang, Zhangyang and Navasardyan, Shant and Shi, Humphrey},
    journal={arXiv preprint arXiv:2403.14773},
    year={2024}
  }

@misc{framepack,
      title={Frame Context Packing and Drift Prevention in Next-Frame-Prediction Video Diffusion Models}, 
      author={Lvmin Zhang and Shengqu Cai and Muyang Li and Gordon Wetzstein and Maneesh Agrawala},
      year={2025},
      eprint={2504.12626},
      archivePrefix={arXiv},
      primaryClass={cs.CV},
      url={https://arxiv.org/abs/2504.12626}, 
}

@misc{pfp,
  title         = {Pretraining Frame Preservation in Autoregressive Video Memory Compression},
  author        = {Lvmin Zhang and Shengqu Cai and Muyang Li and
                   Chong Zeng and Beijia Lu and Anyi Rao and
                   Song Han and Gordon Wetzstein and Maneesh Agrawala},
  year          = {2025},
  eprint        = {2512.23851},
  archivePrefix = {arXiv},
  primaryClass  = {cs.CV},
  url           = {https://arxiv.org/abs/2512.23851}
}

@misc{onestory,
      title={OneStory: Coherent Multi-Shot Video Generation with Adaptive Memory}, 
      author={Zhaochong An and Menglin Jia and Haonan Qiu and Zijian Zhou and Xiaoke Huang and Zhiheng Liu and Weiming Ren and Kumara Kahatapitiya and Ding Liu and Sen He and Chenyang Zhang and Tao Xiang and Fanny Yang and Serge Belongie and Tian Xie},
      year={2025},
      eprint={2512.07802},
      archivePrefix={arXiv},
      primaryClass={cs.CV},
      url={https://arxiv.org/abs/2512.07802}, 
}

@misc{selfforcingpp,
      title={Self-Forcing++: Towards Minute-Scale High-Quality Video Generation}, 
      author={Justin Cui and Jie Wu and Ming Li and Tao Yang and Xiaojie Li and Rui Wang and Andrew Bai and Yuanhao Ban and Cho-Jui Hsieh},
      year={2025},
      eprint={2510.02283},
      archivePrefix={arXiv},
      primaryClass={cs.CV},
      url={https://arxiv.org/abs/2510.02283}, 
}

@misc{rollingforcing,
      title={Rolling Forcing: Autoregressive Long Video Diffusion in Real Time}, 
      author={Kunhao Liu and Wenbo Hu and Jiale Xu and Ying Shan and Shijian Lu},
      year={2025},
      eprint={2509.25161},
      archivePrefix={arXiv},
      primaryClass={cs.CV},
      url={https://arxiv.org/abs/2509.25161}, 
}

@misc{sparseforcing,
      title={Sparse Forcing: Native Trainable Sparse Attention for Real-time Autoregressive Diffusion Video Generation}, 
      author={Boxun Xu and Yuming Du and Zichang Liu and Siyu Yang and Ziyang Jiang and Siqi Yan and Rajasi Saha and Albert Pumarola and Wenchen Wang and Peng Li},
      year={2026},
      eprint={2604.21221},
      archivePrefix={arXiv},
      primaryClass={cs.CV},
      url={https://arxiv.org/abs/2604.21221}, 
}

@inproceedings{storydiffusion,
  title     = {{StoryDiffusion}: Consistent Self-Attention for
               Long-Range Image and Video Generation},
  author    = {Zhou, Yupeng and Zhou, Daquan and Cheng, Ming-Ming
               and Feng, Jiashi and Hou, Qibin},
  booktitle = {Advances in Neural Information Processing Systems},
  year      = {2024}
}

@misc{dreamfactory,
      title={DreamFactory: Pioneering Multi-Scene Long Video Generation with a Multi-Agent Framework}, 
      author={Zhifei Xie and Daniel Tang and Dingwei Tan and Jacques Klein and Tegawend F. Bissyand and Saad Ezzini},
      year={2024},
      eprint={2408.11788},
      archivePrefix={arXiv},
      primaryClass={cs.AI},
      url={https://arxiv.org/abs/2408.11788}, 
}

@misc{movieagent,
      title={Automated Movie Generation via Multi-Agent CoT Planning}, 
      author={Weijia Wu and Zeyu Zhu and Mike Zheng Shou},
      year={2025},
      eprint={2503.07314},
      archivePrefix={arXiv},
      primaryClass={cs.CV},
      url={https://arxiv.org/abs/2503.07314}, 
}

@misc{animaker,
      title={AniMaker: Multi-Agent Animated Storytelling with MCTS-Driven Clip Generation}, 
      author={Haoyuan Shi and Yunxin Li and Xinyu Chen and Longyue Wang and Baotian Hu and Min Zhang},
      year={2025},
      eprint={2506.10540},
      archivePrefix={arXiv},
      primaryClass={cs.MA},
      url={https://arxiv.org/abs/2506.10540}, 
}

@misc{videomemory,
      title={VideoMemory: Toward Consistent Video Generation via Memory Integration}, 
      author={
  Jinsong Zhou and Yihua Du and Xinli Xu and Luozhou Wang and
  Zijie Zhuang and Yehang Zhang and Shuaibo Li and Xiaojun Hu and
  Bolan Su and Chen, Ying-cong
},
      year={2026},
      eprint={2601.03655},
      archivePrefix={arXiv},
      primaryClass={cs.CV},
      url={https://arxiv.org/abs/2601.03655}, 
}

@InProceedings{vbench,
  title={{VBench}: Comprehensive Benchmark Suite for Video Generative Models},
  author={Huang, Ziqi and He, Yinan and Yu, Jiashuo and Zhang, Fan and Si, Chenyang and Jiang, Yuming and Zhang, Yuanhan and Wu, Tianxing and Jin, Qingyang and Chanpaisit, Nattapol and Wang, Yaohui and Chen, Xinyuan and Wang, Limin and Lin, Dahua and Qiao, Yu and Liu, Ziwei},
  booktitle={Proceedings of the IEEE/CVF Conference on Computer Vision and Pattern Recognition},
  year={2024}
}

@article{vbench2,
      title={{VBench-2.0}: Advancing Video Generation Benchmark Suite for Intrinsic Faithfulness},
      author={Zheng, Dian and Huang, Ziqi and Liu, Hongbo and Zou, Kai and He, Yinan and Zhang, Fan and Zhang, Yuanhan and He, Jingwen and Zheng, Wei-Shi and Qiao, Yu and Liu, Ziwei},
      journal={arXiv preprint arXiv:2503.21755},
      year={2025}
}

@misc{vistorybench,
      title={ViStoryBench: Comprehensive Benchmark Suite for Story Visualization}, 
      author={Cailin Zhuang and Ailin Huang and Yaoqi Hu and Jingwei Wu and Wei Cheng and Jiaqi Liao and Hongyuan Wang and Xinyao Liao and Weiwei Cai and Hengyuan Xu and Xuanyang Zhang and Xianfang Zeng and Zhewei Huang and Gang Yu and Chi Zhang},
      year={2026},
      eprint={2505.24862},
      archivePrefix={arXiv},
      primaryClass={cs.CV},
      url={https://arxiv.org/abs/2505.24862}, 
}

@misc{IAMFlow,
      title={Advancing Narrative Long Video Generation via Training-Free Identity-Aware Memory}, 
      author={Jinzhuo Liu and Jiangning Zhang and Wencan Jiang and Yabiao Wang and Dingkang Liang and Zhucun Xue and Ran Yi and Yong Liu},
      year={2026},
      eprint={2605.18733},
      archivePrefix={arXiv},
      primaryClass={cs.CV}
}

@misc{gpt41,
  author       = {{OpenAI}},
  title        = {Introducing GPT-4.1 in the API},
  year         = {2025},
  month        = apr,
  howpublished = {\url{https://openai.com/index/gpt-4-1/}}
}

@misc{gemini3pro,
  author       = {{Google DeepMind}},
  title        = {Gemini 3 Pro Model Card},
  year         = {2026},
  month        = may,
  howpublished = {\url{https://deepmind.google/models/model-cards/gemini-3-pro}},
  note         = {Model released in November 2025; model card updated in May 2026}
}
\bibliographystyle{plainnat}
\newpage
\clearpage
\appendix
\section{Inference Phase Pipeline}
\begin{algorithm}[H]
\caption{Inference of SlotMem}
\label{alg:Slotmem_inference}
\begin{algorithmic}[1]
\Require Chunk prompts $\mathbf{P}=[\mathbf{p}^{1},\ldots,\mathbf{p}^{K}]$,
initial image condition $\mathbf{I}^{0}$, denoising timesteps
$\{t_s\}_{s=1}^{S}$, selected layer set $\mathcal{L}_{p}$
used for probing, memory extraction, and injection,
layer-group map $g(\ell_p)$ for $\ell_p\in\mathcal{L}_{p}$,
group-specific memory encoders $\{E_g\}$,
Memory Writer $\operatorname{Writer}$
\Ensure Generated chunks $\mathbf{X}=[\mathbf{x}^{1},\ldots,\mathbf{x}^{K}]$

\State Initialize role-wise slot memory bank $\mathcal{B}\leftarrow\emptyset$
and image condition $\mathbf{I}^{1}\leftarrow\mathbf{I}^{0}$

\For{$k=1$ to $K$}
    \State Extract character set $\mathcal{C}^{k}$ from prompt $\mathbf{p}^{k}$
    and retrieve the corresponding layer-wise slots $\mathcal{R}^{k}$ from
    $\mathcal{B}$
    \State Sample initial noisy latent $\mathbf{z}^{k}_{t_1}$
    \State Initialize cached role masks $\bar{\mathcal{P}}^{k}\leftarrow\emptyset$
    and online slot buffer $\mathcal{U}^{k}\leftarrow\emptyset$

    \For{$s=1$ to $S$}
                \State
        $\begin{aligned}[t]
        &(\mathbf{z}^{k}_{t_{s+1}},\mathcal{P}^{k}_{s},\mathcal{V}^{k}_{s})
        \leftarrow \textproc{MemoryAwareDiTForward}( \\
        &\qquad \mathbf{z}^{k}_{t_s},\mathbf{I}^{k},\mathbf{p}^{k},\mathcal{R}^{k},\bar{\mathcal{P}}^{k},\mathcal{L}_{p},g,\{E_g\})
        \end{aligned}$
        \State Cache current masks for the next step:
        $\bar{\mathcal{P}}^{k}\leftarrow\mathcal{P}^{k}_{s}$
        \State Accumulate current online slots:
        $\mathcal{U}^{k}\leftarrow\mathcal{U}^{k}\cup\mathcal{V}^{k}_{s}$
    \EndFor

    \State Decode $\mathbf{z}^{k}_{t_{S+1}}$ to obtain $\mathbf{x}^{k}$
    \State Insert new slots into $\mathcal{B}$ or update existing slots with
    $\mathcal{U}^{k}$ using the Memory Writer according to
    Eqs.~\eqref{mem up delta}--\eqref{mem up}
    \State Update $\mathbf{I}^{k+1}$ using the ending frame of $\mathbf{x}^{k}$
\EndFor

\State \Return $\mathbf{X}$

\vspace{2mm}
\Procedure{MemoryAwareDiTForward}{}(
\Statex \hspace{\algorithmicindent}
$\begin{aligned}[t]
&\mathbf{z}^{k}_{t_s},\mathbf{I}^{k},\mathbf{p}^{k},
\mathcal{R}^{k},\bar{\mathcal{P}}^{k},\mathcal{L}_{p},g,\{E_g\})
\end{aligned}$

\State Run one DiT denoising forward on $\mathbf{z}^{k}_{t_s}$ with condition
$(\mathbf{I}^{k},\mathbf{p}^{k})$, and cache cross-attention responses and
hidden states at layers $\ell_p\in\mathcal{L}_{p}$
\If{$\mathcal{R}^{k}\neq\emptyset$ and
$\bar{\mathcal{P}}^{k}\neq\emptyset$}
    \State During the forward, inject retrieved slots at layers
    $\ell_p\in\mathcal{L}_{p}$ using
    Character-Wise Cross-Attention according to
    Eqs.~\eqref{mem in probe}--\eqref{eq:sparse_writeback}
\EndIf
\State Predict the denoising update and obtain $\mathbf{z}^{k}_{t_{s+1}}$
\State Compute current role masks $\mathcal{P}^{k}_{s}$ with
Character-Semantic Probe according to
Eqs.~\eqref{eq:char_response}--\eqref{eq:char_mask}
\State Extract role tokens at layers $\ell_p\in\mathcal{L}_{p}$ and encode
online slots $\mathcal{V}^{k}_{s}$ with the memory encoder $E_{g(\ell_p)}$
according to Eqs.~\eqref{eq:Mem exe probe}--\eqref{mem exe encode}

\State \Return
$(\mathbf{z}^{k}_{t_{s+1}},
\mathcal{P}^{k}_{s},\mathcal{V}^{k}_{s})$
\EndProcedure

\end{algorithmic}
\end{algorithm}
\section{Additional Experimental Settings}
\label{extra_setting}

\paragraph{Baselines.}
We compare our method with the base model Wan2.2-I2V under the autoregressive video generation setting and three representative baselines implemented on the same base model. StoryDiffusion~\cite{storydiffusion} is a classical baseline that first generates consistent multi-image references and then uses them as the initial-frame conditions for different chunks; in our implementation, we use Wan2.2-I2V as its video diffusion backbone for chunk generation. StoryMem~\cite{storymem} is an advanced memory-based method that retrieves memory frames based on their aesthetic scores and their diversity with respect to the existing memory bank; we directly follow its original Wan2.2-based setting. IAMFlow~\cite{IAMFlow} is a training-free identity-aware memory framework that uses VLM agents to maintain persistent entity IDs and retrieve identity-relevant memory frames across prompts. Since its released implementation is based on Wan2.1-T2V, we adapt it to Wan2.2-I2V while keeping its default hyperparameters for memory unchanged, such as the number of memory frames maintained for each character.

\begin{figure*}[t]
 \centering
\includegraphics[width=\textwidth]{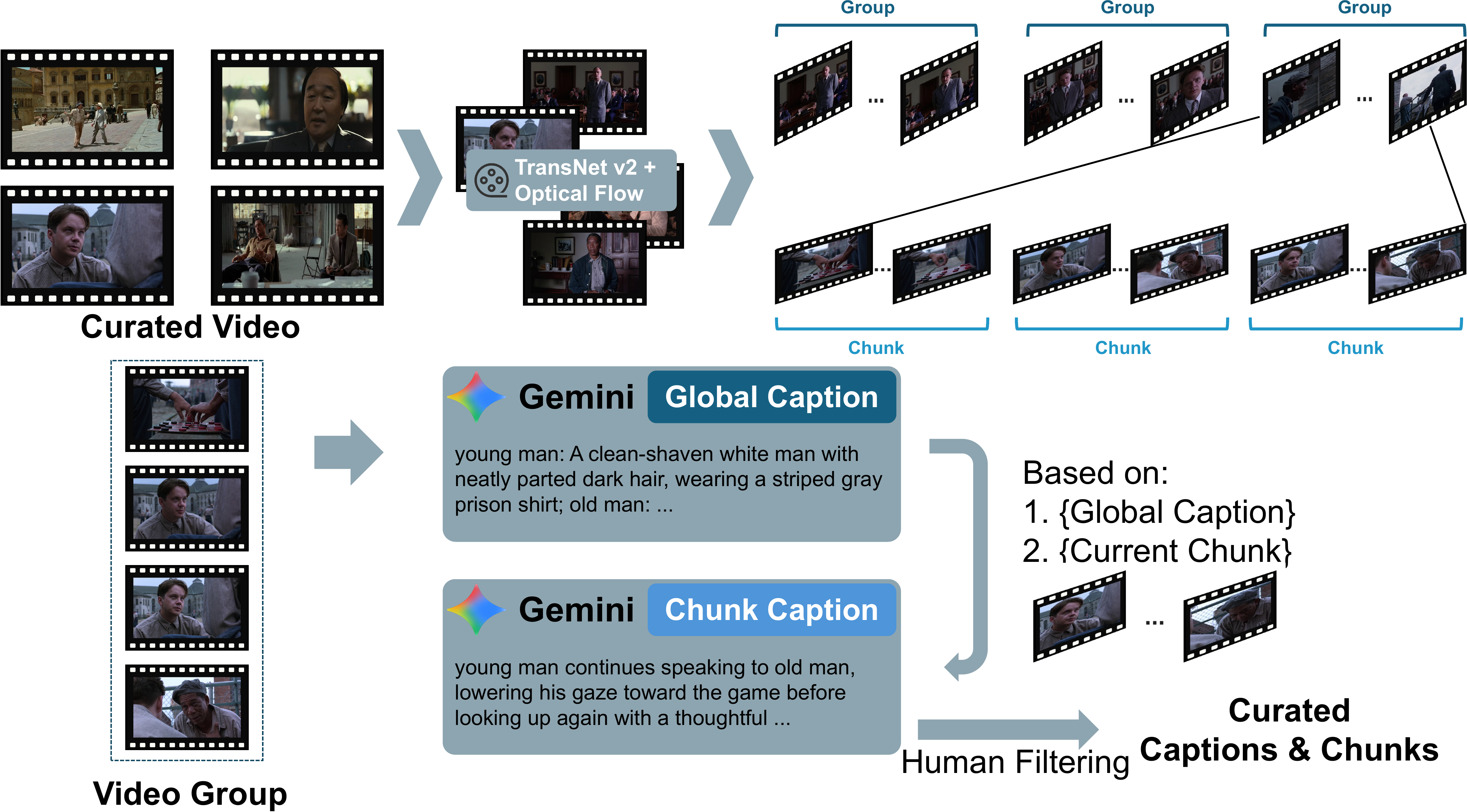}
\caption{Data Curation Pipeline, adopted from MultiShotMaster. The Subject Merging component is removed since our task is not subject-to-video generation.}
\label{fig:curation_pipe}
\end{figure*}

\paragraph{Data Curation}
We construct the training dataset using a multi-stage curation pipeline adapted from MultiShotMaster~\cite{multishotmaster}, as illustrated in Fig.~\ref{fig:curation_pipe}. Source videos are first normalized to 24 FPS, after which TransNetV2 and optical flow are applied to detect shot boundaries. Each retained group is divided into 81-frame chunks. Gemini 3.0~\cite{gemini3pro} first generates a global caption from uniformly sampled frames within each shot group, defining consistent identifiers for the main characters and summarizing the scene and visual style. Conditioned on this global caption, the model then produces a caption for each chunk that describes character actions and expressions, the scene background, and camera motion. Finally, character mentions are extracted from the captions to construct chunk-level character lists and cross-chunk associations, which are used to identify valid training samples and establish the corresponding core--memory relationships.

\paragraph{Dataset Availability and Copyright.}
Our training and evaluation data are derived from publicly accessible films and online videos. However, public accessibility does not imply redistribution rights. Since the underlying video content is copyrighted and the authors do not own the rights required to redistribute the processed clips, we cannot release the video data under a research-use license.
\begin{figure}[!ht]
\centering

% Left: test dataset distribution
\begin{minipage}[c]{0.32\linewidth}
  \centering
  \includegraphics[width=\linewidth]{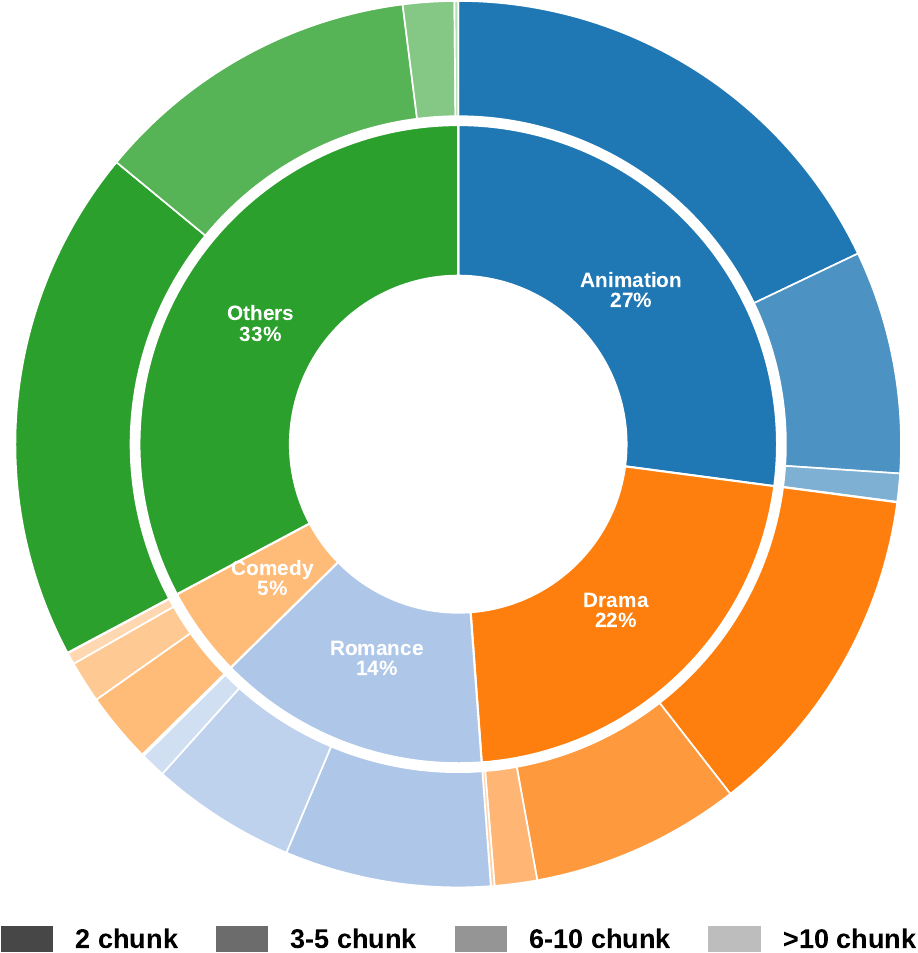}
\end{minipage}
\hfill
% Middle: character composition
\begin{minipage}[c]{0.32\linewidth}
  \centering
  \includegraphics[width=\linewidth]{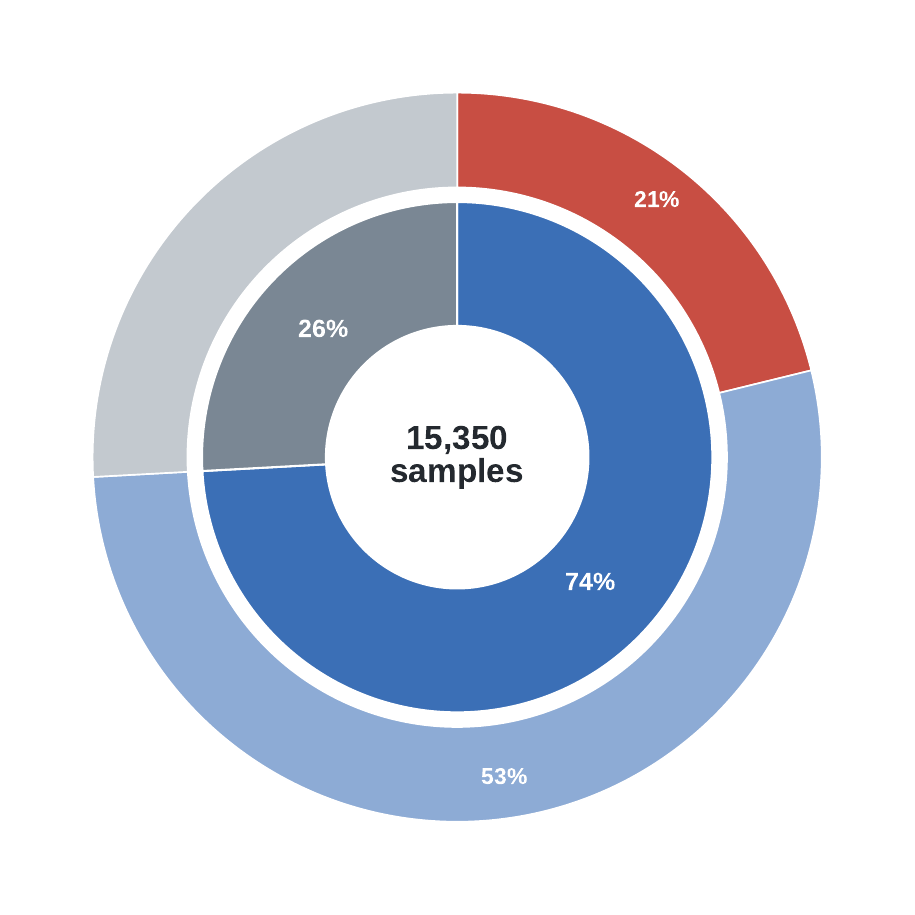}
\end{minipage}
\hfill
% Right: legend and training dataset statistics
\begin{minipage}[c]{0.32\linewidth}
  \centering
  \includegraphics[width=0.92\linewidth]{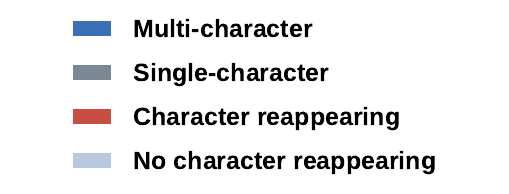}

  \vspace{0.02\linewidth}

  \small
  \renewcommand{\arraystretch}{1.12}
  \begin{tabular*}{\linewidth}{@{}l@{\extracolsep{\fill}}r@{}}
    \toprule
    Dataset Statistic & \\
    \midrule
    Number of samples & 15,350 \\
    Number of chunks  & 42,320 \\
    Duration          & 39.71 h \\
    \bottomrule
  \end{tabular*}
\end{minipage}

\caption{
Dataset statistics. We report the category and chunk-count distribution of the test samples (left), the character composition of the training dataset (middle), and its legend and basic statistics (right).
}
\label{fig:train_dataset_overview}

\end{figure}
\paragraph{Dataset Details}
Based on the curated samples, we categorize the dataset according to character composition and reappearance patterns. Specifically, samples are divided into single-character samples and multi-character samples; within the multi-character subset, we further distinguish samples with character reappearance from those without character reappearance. Here, character reappearance refers to the case where a character appears in an earlier chunk, disappears in one or more subsequent chunks, and then appears again in a later chunk. The character composition and dataset statistics are summarized in Fig.~\ref{fig:train_dataset_overview}.

For evaluation, we randomly select 60 samples from additional video sources while matching the character-composition ratio of the training dataset. The test set therefore follows the same distribution over single-character samples, multi-character samples without character reappearing, and multi-character samples with character reappearing, while using video sources disjoint from the training set.

\paragraph{Additional Training Details}

We train the model on NVIDIA A100 GPUs for one epoch with 480 GPU-hours using DDP. The learning rate is set to \(1\times10^{-4}\), and the noise-domain boundary ratio in WAN2.2 is set to \(0.9\). For the memory module, we use DiT layers \(0\text{--}15\) for role-token extraction and sparse memory injection. These layers are divided into three groups, \(0\text{--}4\), \(5\text{--}10\), and \(11\text{--}15\), each equipped with an independent Memory Encoder and Writer. The sparse role memory uses 8 heads with a head dimension of 128, a RoPE dimension of 256. The memory top-\(p\) ratio is set to \(0.1\) and \(\lambda_{\mathrm{slot}}\) is set to 0.02.

\begin{figure}[!ht] 
 \centering
\includegraphics[width=\textwidth]{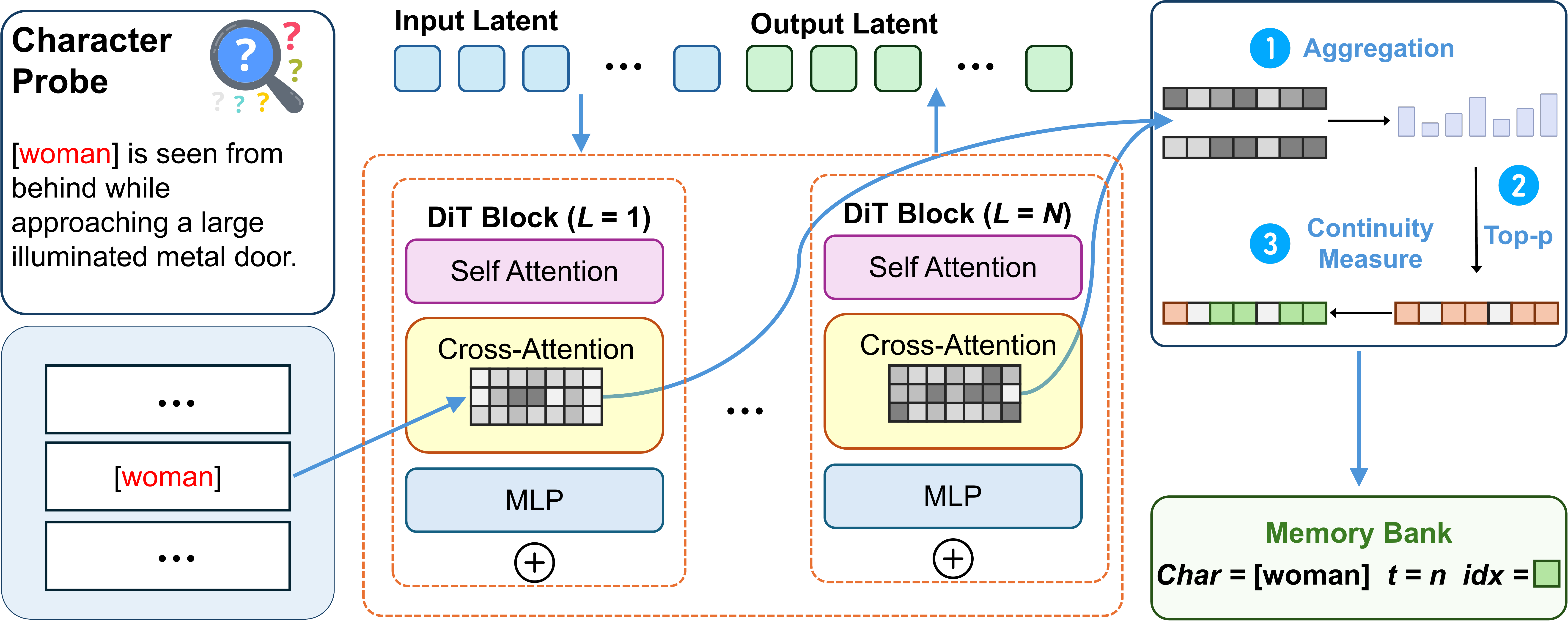}
  \caption{\textbf{Character-Semantic Probe.} Given an input latent, the probe aggregates cross-attention responses to each character name, computes character-semantic responses, selects top-\(p\) character-sensitive tokens, and filters isolated noisy tokens for memory construction and sparse injection.}
 \label{fig:probe}
\end{figure}

\begin{figure*}[!ht] % [!ht] 是一个可选参数,让LaTeX尽量把图片放在这里 (Here) 或顶部 (Top)
 \centering
\includegraphics[width=\textwidth]{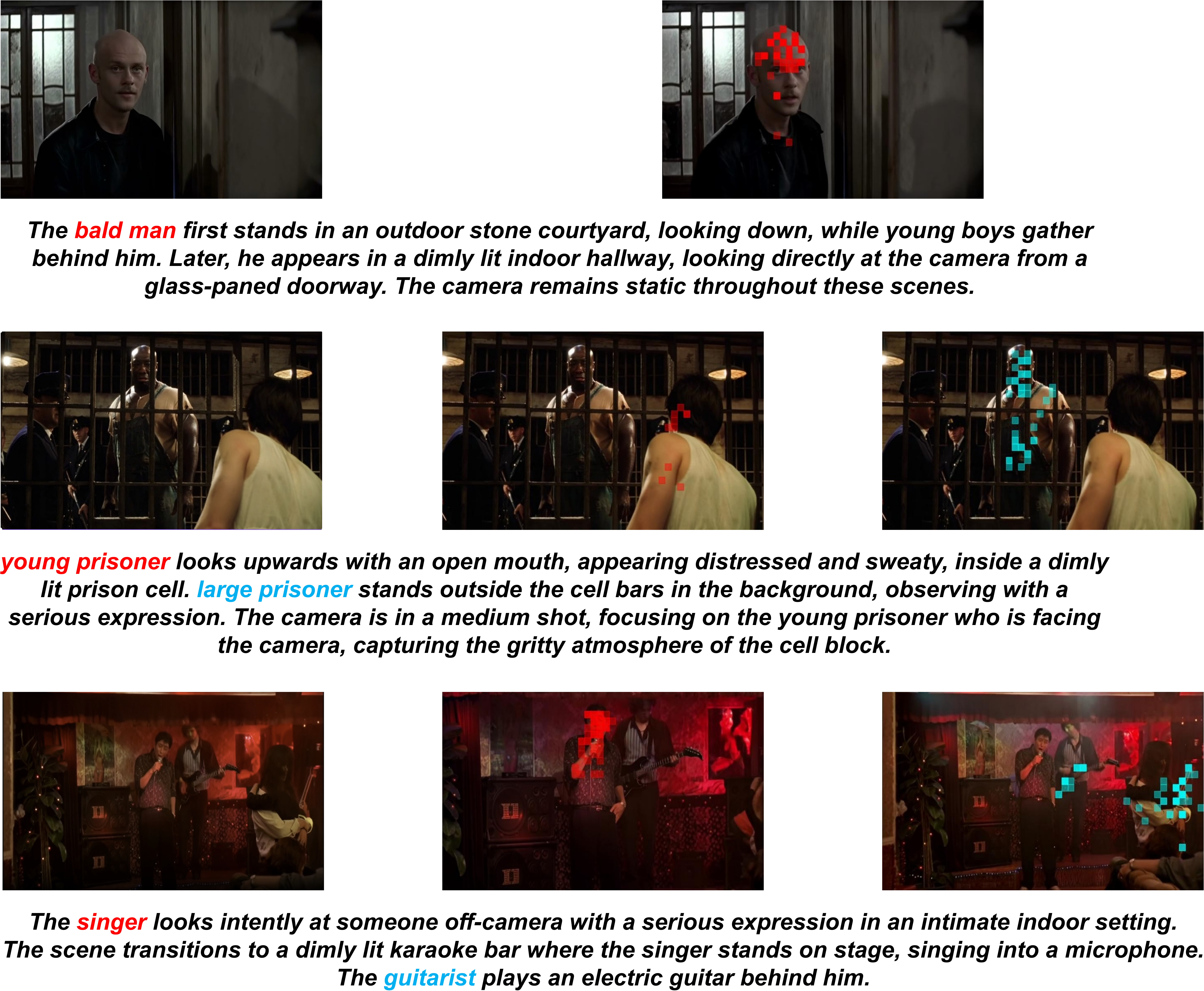}
\caption{
Visualization of Character-Semantic Probe Regions in Pixel Space. We remap the latent tokens selected by the Character-Semantic Probe to the original pixel space and visualize the first frame of each corresponding 8-frame patch. For visualization, the colored character token in each prompt is matched with the memory patch of the same color shown above.
}
 \label{fig:memory_viz} % 整个图的引用标签
\end{figure*}

\begin{figure*}[t]
 \centering
\includegraphics[width=\textwidth]{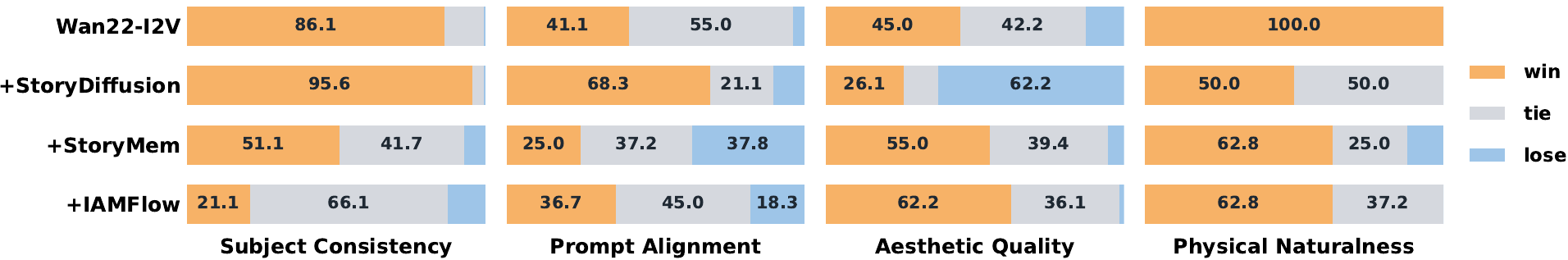}
\caption{Human Preference Study.}
\label{fig:human_eval}
\includegraphics[width=\textwidth]{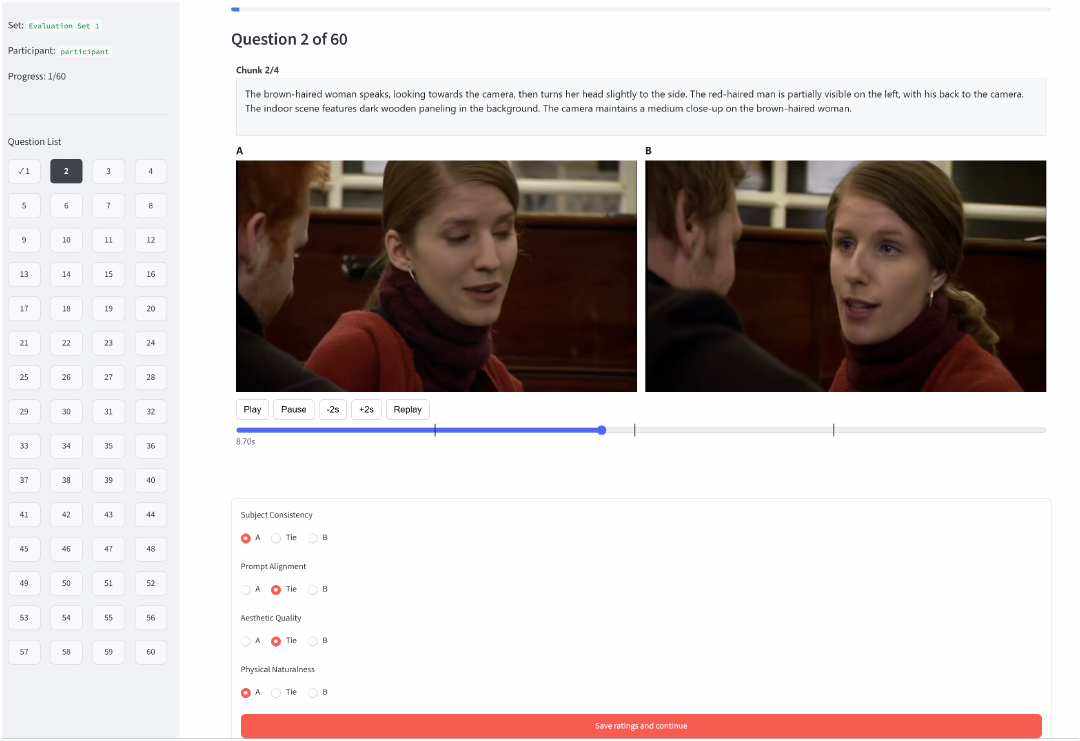}
\caption{The interface of the Human Preference Study.}
\label{fig:human_interface}
\vspace{-1.9em}
\end{figure*}

\section{Visualization of Character-Semantic Probe Region in Pixel Space}

Although the Character-Semantic Probe, as illustrated in Fig.~\ref{fig:probe}, localizes character-sensitive positions in the latent space, its effectiveness can also be examined directly in the pixel space. This is because the VAE, unlike a semantic encoder, retains a relatively direct spatial correspondence between pixel-space regions and latent-space tokens. For visualization, we map the selected latent tokens back to the original pixel space during inference and display the first frame of each corresponding 8-frame patch.

As illustrated in Fig.~\ref{fig:memory_viz}, the Character-Semantic Probe localizes character-relevant regions in both single-character and multi-character scenes, and it successfully distinguishes different characters in the two examples in the top row. However, when the automatic captioning pipeline fails to capture the distinctive attributes of a character within a given chunk, the resulting prompt ambiguity may lead to mixed probe regions, as observed for the guitarist in the last example. This confusion may subsequently affect the quality of memory extraction and injection.

\begin{table*}[!ht]
\centering
\scriptsize
\setlength{\tabcolsep}{3.0pt}
\renewcommand{\arraystretch}{1.04}
\setlength{\extrarowheight}{0pt}

\setlength{\aboverulesep}{0pt}
\setlength{\belowrulesep}{0pt}
\setlength{\cmidrulesep}{0pt}

\begin{tabularx}{\linewidth}{@{}>{\raggedright\arraybackslash}p{0.205\textwidth}>{\raggedright\arraybackslash}p{0.125\textwidth}*{5}{Y}@{}}
\toprule
\textbf{Metric} & \textbf{VLM}
& \textbf{Wan2.2}
& {\scriptsize\textbf{StoryDiffusion}}
& \textbf{StoryMem}
& \textbf{IAMFlow}
& \textbf{SlotMem} \\
\midrule
\multicolumn{7}{c}{\textbf{ViStoryBench}} \\
\midrule
Prompt Alignment & Qwen3.5-4B & \underline{0.9255} & 0.9035 & \third{0.9172} & 0.8793 & \textbf{0.9326} \\
 & Qwen3-VL-30B & \underline{0.8302} & 0.8264 & \third{0.8288} & 0.7944 & \textbf{0.8542} \\
 & GPT-4.1 & \underline{0.8299} & 0.7627 & \third{0.8273} & 0.7192 & \textbf{0.8733} \\
\midrule
\multicolumn{7}{c}{\textbf{NarraStream-Bench}} \\
\midrule
Conditional Adjacent & Qwen3.5-4B & 0.3785 & 0.3322 & \third{0.4525} & \textbf{0.6176} & \underline{0.6159} \\
 & Qwen3-VL-30B & 0.3632 & 0.3245 & \third{0.4693} & \textbf{0.6376} & \underline{0.6126} \\
 & GPT-4.1 & 0.3632 & 0.3264 & \third{0.4556} & \textbf{0.6131} & \underline{0.6126} \\
Conditional Long-range & Qwen3.5-4B & \underline{0.7200} & 0.5800 & 0.6147 & \third{0.7172} & \textbf{0.8416} \\
 & Qwen3-VL-30B & \third{0.7200} & 0.5774 & 0.6158 & \underline{0.7289} & \textbf{0.8416} \\
 & GPT-4.1 & \third{0.7239} & 0.5880 & 0.6165 & \underline{0.7313} & \textbf{0.8363} \\
Entity Grounding & Qwen3.5-4B & 0.8799 & \third{0.9059} & \underline{0.9104} & 0.9027 & \textbf{0.9206} \\
 & Qwen3-VL-30B & \textbf{0.7993} & \third{0.7711} & \underline{0.7792} & 0.7605 & 0.7504 \\
 & GPT-4.1 & \third{0.6633} & 0.5675 & \underline{0.6648} & 0.6029 & \textbf{0.6735} \\
VLM Score & Qwen3.5-4B & \underline{0.9367} & \third{0.9272} & 0.7118 & 0.9174 & \textbf{0.9614} \\
 & Qwen3-VL-30B & \third{0.8198} & 0.6253 & \textbf{0.8318} & 0.7201 & \underline{0.8291} \\
 & GPT-4.1 & \third{0.4708} & 0.2817 & \underline{0.4735} & 0.4607 & \textbf{0.5384} \\
\bottomrule
\end{tabularx}

\caption{
VLM-sensitive metrics under different VLM judges. Prompt Alignment is from ViStoryBench, while Conditional Adjacent, Conditional Long-range, Entity Grounding, and VLM Score are from NarraStream-Bench. SlotMem achieves the best performance on nearly all VLM metrics across different VLM judges.
}
\label{fig:vlm_judge_sensitive_metrics}
\end{table*}

\section{Human Preference Study}
Although standard benchmarks provide comprehensive automatic evaluation, their metrics cannot fully reflect perceptual factors such as identity preservation and natural cross-shot transitions. Therefore, we additionally conduct a human preference study on the test samples. For each comparison, evaluators is shown the prompt and two anonymized generated videos, one from SlotMem and one from a baseline, with the A/B order shuffled to reduce positional bias. The evaluator then selects whether SlotMem wins, loses, or ties with the baseline along four dimensions: subject consistency, prompt alignment, aesthetic quality, and physical naturalness. The evaluation interface is illustrated in Fig.~\ref{fig:human_interface}. We aggregate the results as win/tie/lose percentages for SlotMem against each baseline. As shown in Fig.~\ref{fig:human_eval}, SlotMem receives higher user preference than the competing methods in most evaluation dimensions, with particularly clear advantages in subject consistency and physical naturalness, which are closely related to character identity preservation and realistic character dynamics.
\begin{table}[!ht]
\centering
\caption{Efficiency comparison across baselines and SlotMem with full memory and condition buffers under standard hyperparameters.}
\label{tab:strict_full_buffer_efficiency}
\resizebox{\linewidth}{!}{%
\begin{tabular}{lcccc}
\toprule
Method & GPU Memory (GB) & Latency (s) & FPS & PFLOPs \\
\midrule
\multicolumn{5}{l}{\textit{Backbone}} \\
Wan2.2 i2v & 32.89 & 993.26 & 0.0816 & 78.59 \\
\midrule
\multicolumn{5}{l}{\textit{Training-free baselines}} \\
StoryDiffusion & 32.89 & 1000.53 & 0.0810 & 78.86 \\
IAMFlow & 32.92 & 972.73 & 0.0833 & 78.59 \\
\midrule
\multicolumn{5}{l}{\textit{Fine-tuned methods}} \\
StoryMem & 78.43 & 2042.09 & 0.0392 & 99.48 \\
SlotMem & 37.24 & 1786.62 & 0.0453 & 84.92 \\
\bottomrule
\end{tabular}%
}
\vspace{0.25em}
\end{table}

% \begin{figure}[H]
% \centering
% \includegraphics[width=0.9\linewidth,height=0.25\textheight,keepaspectratio]{radars.pdf}

% \caption{Radar visualization.}
% \label{fig:radar}
% \end{figure}

\section{Limitations}
\label{limitation}
Despite these promising results, our current framework still has several limitations.
\paragraph{Training Scale.}
Due to limited training scale, our model has not yet been fully evaluated under larger and more diverse training regimes. Although the current results demonstrate the effectiveness of our framework, we expect that scaling up the training data, increasing style diversity, and incorporating videos from broader domains would further improve its robustness and generalization. More importantly, such scaling would provide a stronger validation of our method under more realistic and large-scale multi-shot video generation settings.

\paragraph{Dependence on Character-Semantic Anchors.}
Our framework relies on reliable character-semantic anchors in the chunk-level prompts. This places a relatively strict requirement on the automatic captioning pipeline: recurring characters should be named consistently, and incorrect or ambiguous character annotations may affect both memory construction and memory injection, just like \textit{[guitarist]} in the last sample. This issue becomes more challenging when multiple visually similar identities appear in the same frame, where simple textual character names may not be sufficient for the probe to extract fully independent feature memory for each role. A promising future direction is to leverage caption data with more fine-grained character descriptions, and extend our overall pipeline by replacing simple character-name anchors with richer attribute-level character anchors. Such fine-grained anchors could provide more precise guidance for memory localization and injection, thereby improving identity disambiguation in challenging multi-character scenes.

\begin{figure*}[t]
 \centering
\includegraphics[width=\textwidth]{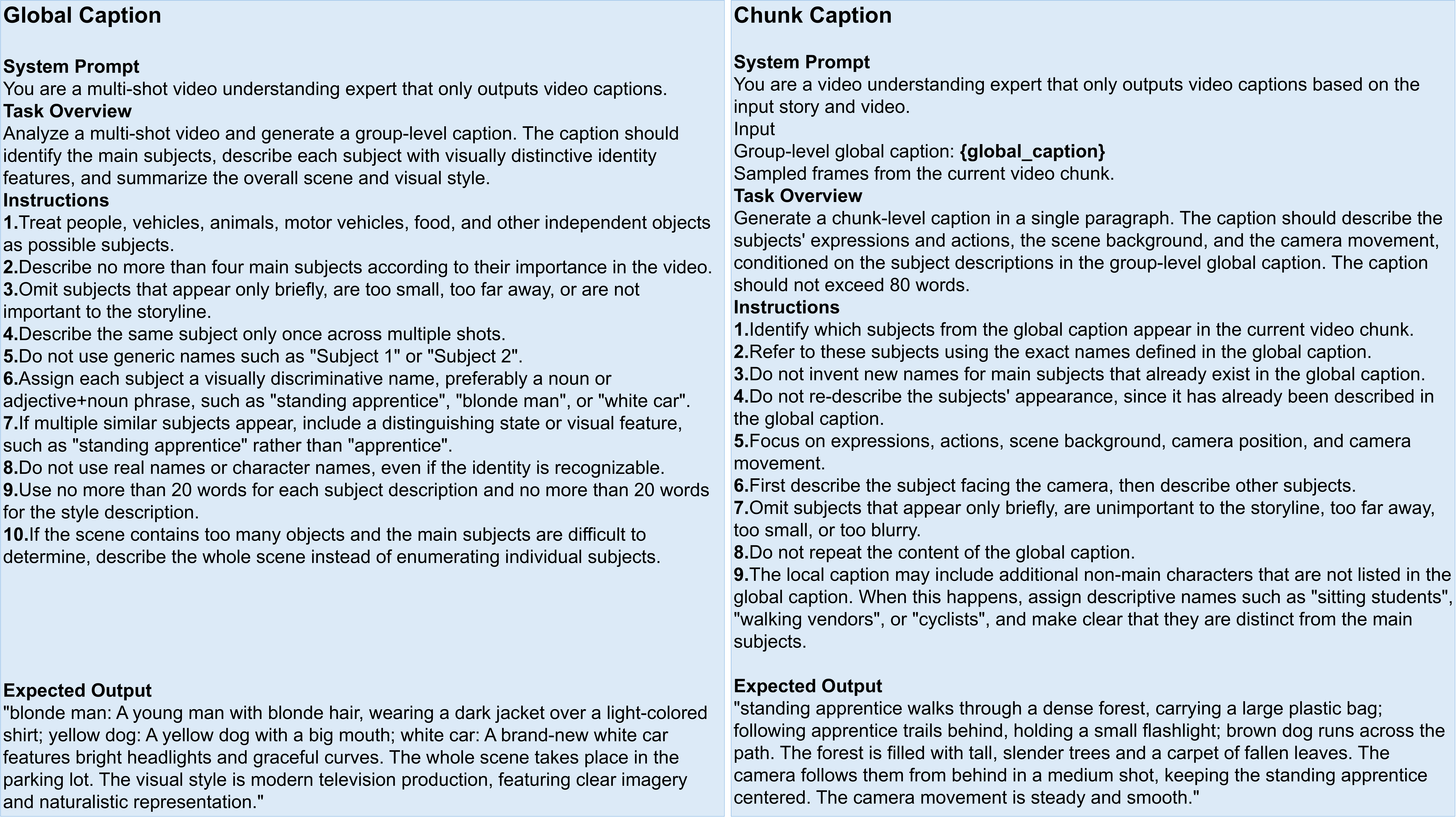}
\caption{Prompts used for hierarchical caption annotation of the curated data. The global caption prompt (left) is used to build a consistent story-level description across the whole video, while the chunk caption prompt (right) is used to generate captions for individual chunks. Rules and Expected output are redesigned based on the requirement for Character-Semantic Probe}
\label{fig:prompt}
\includegraphics[width=\textwidth]{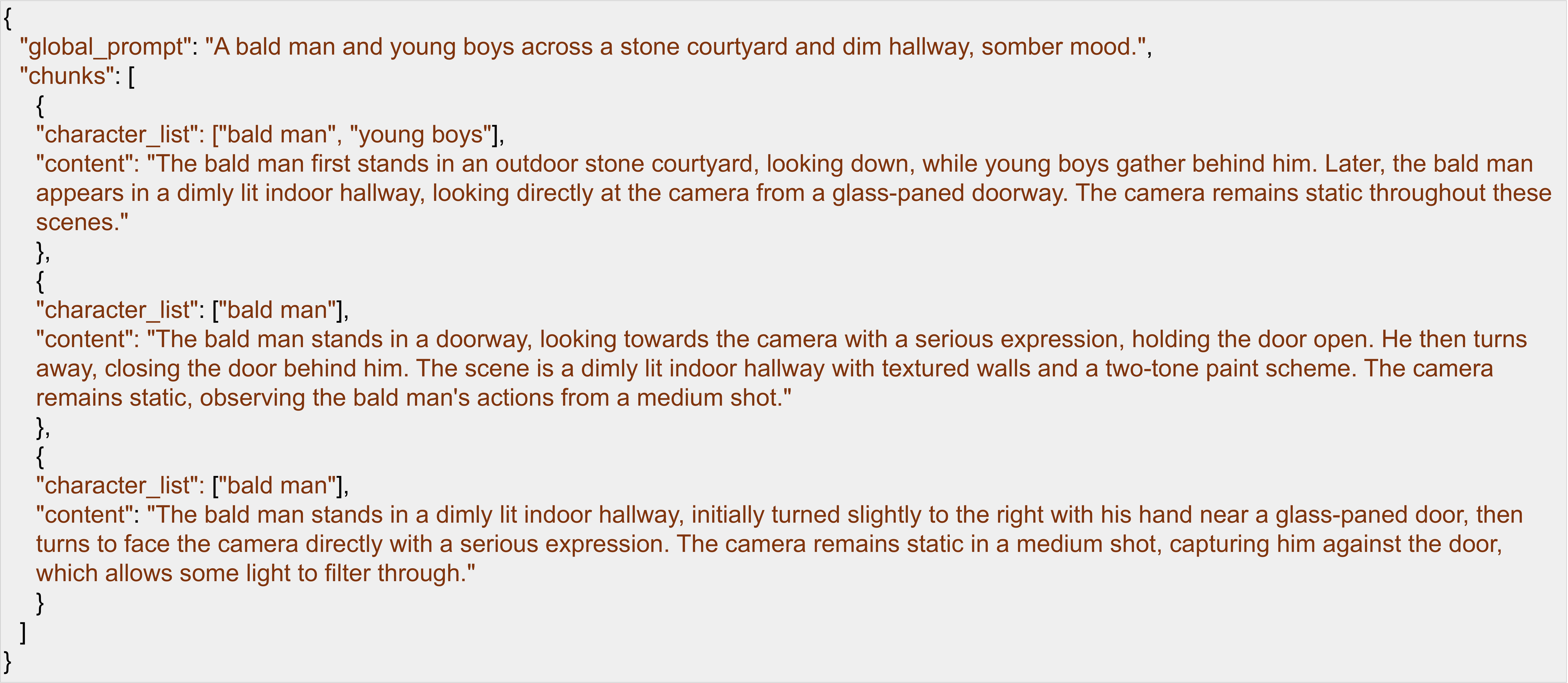}
\caption{Example of the story caption.}
\label{fig:sample}
\end{figure*}
%%%%%%%%%%%%%%%%%%%%%%%%%%%%%%%%%%%%%%%%%%%%%%%%%%%%%%%%%%%%
% \section{Broader Impact.}
% \label{Impact}
% SlotMem aims to improve controllable narrative long video generation by reducing long-range identity drift in multi-character stories. This can benefit creative applications such as storyboarding, animation prototyping, and assistive content creation, where maintaining character identity across long videos is important. At the same time, improving identity consistency and temporal coherence may also increase the realism and credibility of synthetic videos, which could be misused for deceptive media generation, impersonation, or other identity-related manipulation. In addition, our training data are collected from publicly available movies and TV series, and we do not redistribute them to respect copyright and licensing constraints. We encourage responsible use of the proposed method and recommend that generated content be clearly disclosed as synthetic when deployed in real-world applications.

\end{document}